\documentclass[lettersize,journal]{IEEEtran}
\usepackage{amsmath,amsfonts}
\usepackage{algpseudocode}
\usepackage{algorithm}
\usepackage{array}
\usepackage{subcaption}
\usepackage{textcomp}
\usepackage{stfloats}
\usepackage{url}
\usepackage{verbatim}
\usepackage{graphicx}
\usepackage{cite}
\usepackage{hyperref}
\usepackage{adjustbox} % For tables with specified width
\usepackage{booktabs}
\usepackage{multirow}
\usepackage[table,xcdraw]{xcolor}
\usepackage{xcolor}
\usepackage{pifont}
\usepackage{placeins} % Load the placeins package for \FloatBarrier
\usepackage{cite}
\usepackage{tikz}
\usetikzlibrary{shapes.multipart}
\usetikzlibrary{shapes,arrows}
\usepackage{amsmath}
\usepackage{bm}

\begin{document}

\title{Radar-Camera Fused Multi-Object Tracking: Online Calibration and Common Feature}
%Fusing Radar and Camera Data for Multi-Object Tracking: An Online Calibration and Radar-Camera Common Feature-Based Approach

\author{Lei Cheng,~\IEEEmembership{Graduate Student Member,~IEEE}, and Siyang Cao,~\IEEEmembership{Senior Member,~IEEE}
        % <-this % stops a space
\thanks{The authors are with the Department of Electrical and Computer Engineering, The University of Arizona, Tucson, AZ 85721 USA (e-mail: leicheng@arizona.edu; caos@arizona.edu)}% <-this % stops a space
% \thanks{Manuscript received April 19, 2021; revised August 16, 2021.}
}

% The paper headers
\markboth{Journal of \LaTeX\ Class Files,~Vol.X, No.X, X}%
{Shell \MakeLowercase{\textit{et al.}}: A Sample Article Using IEEEtran.cls for IEEE Journals}

%\IEEEpubid{0000--0000/00\$00.00~\copyright~2021 IEEE}
% Remember, if you use this you must call \IEEEpubidadjcol in the second
% column for its text to clear the IEEEpubid mark.

\maketitle

\begin{abstract}
This paper presents a Multi-Object Tracking (MOT) framework that fuses radar and camera data to enhance tracking efficiency while minimizing manual interventions. Contrary to many studies that underutilize radar and assign it a supplementary role—despite its capability to provide accurate range/depth information of targets in a world 3D coordinate system—our approach positions radar in a crucial role. Meanwhile, this paper utilizes common features to enable online calibration to autonomously associate detections from radar and camera. The main contributions of this work include: (1) the development of a radar-camera fusion MOT framework that exploits online radar-camera calibration to simplify the integration of detection results from these two sensors, (2) the utilization of common features between radar and camera data to accurately derive real-world positions of detected objects, and (3) the adoption of feature matching and category-consistency checking to surpass the limitations of mere position matching in enhancing sensor association accuracy. To the best of our knowledge, we are the first to investigate the integration of radar-camera common features and their use in online calibration for achieving MOT. The efficacy of our framework is demonstrated by its ability to streamline the radar-camera mapping process and improve tracking precision, as evidenced by real-world experiments conducted in both controlled environments and actual traffic scenarios.  Code is available at \href{https://github.com/radar-lab/Radar_Camera_MOT}{https://github.com/radar-lab/Radar\_Camera\_MOT}

\end{abstract}


\section{Introduction}
\IEEEPARstart{M}{ulti}-Object Tracking represents a vital research area that has far-reaching implications for various applications, particularly in the burgeoning field of autonomous driving \cite{zhang2023attentiontrack} and intelligent transportation systems \cite{shen2023interactively}. MOT involves identifying and tracking multiple objects over time, providing valuable insights into their motion patterns \cite{cui2023online,leimot}. By continuously monitoring and tracking these objects, the autonomous driving system can anticipate potential risks and adapt its behavior accordingly, ensuring safety and efficiency on the road \cite{tang2021road}. Similarly, by perceiving and predicting the future movements of traffic participants, intelligent transportation systems can enable early collision warnings and proactive risk alerts \cite{zhang2023attentiontrack}, as well as anticipate traffic congestion in advance. These capabilities help protect vulnerable road users and improve overall traffic flow. Most existing MOT systems rely excessively on visual data from cameras and lidar, which can lead to difficulties in complex scenarios, such as occlusions, sudden appearance changes, lighting variations, and adverse weather \cite{luo2021multiple}. To overcome these challenges, integrating data from diverse sensors, like radar and cameras, is a promising solution. Radar and camera are complementary sensing modalities widely exploited in applications like autonomous driving and robotics. Radar can provide accurate range, velocity, and angle information irrespective of illumination and weather conditions \cite{harlow2023new}, while cameras can capture high-resolution visual details \cite{cheng2023online}. By leveraging both sensor modalities, MOT systems can achieve more reliable and robust tracking in diverse environments. 

Traditionally, in the radar-camera fusion-based MOT system, radar point clouds and camera images are used and processed through three distinctive branches \cite{leimot,sengupta2022robust}. The radar branch uses clustering algorithms to detect targets, positioning each at its cluster’s center \cite{deng2022robust}. Concurrently, the camera branch utilizes object detectors to identify objects, pinpointing their positions at the bounding boxes' lower center points, which are then transformed into real-world coordinates. Finally, the sensor fusion branch integrates the real-world positions observed by both radar and camera to obtain complementary perception, thereby bolstering the overall accuracy and robustness of the MOT system. 
\begin{table}[]
\centering
\caption{ERRORS CAUSED BY A PITCH ANGLE CHANGE OF ±0.5 DEGREES AND A HEIGHT CHANGE OF ±0.05 METERS DURING BEV TRANSFORMATION}
\label{tabbev}
\resizebox{0.47\textwidth}{!}{%
\begin{tabular}{c|ccc|ccc|}
\cline{2-7}

 &
  \multicolumn{3}{c|}{\cellcolor[HTML]{FFFFFF}\begin{tabular}[c]{@{}c@{}} \textbf{Pitch Angle}\\ (degrees)\end{tabular}} &
  \multicolumn{3}{c|}{\cellcolor[HTML]{FFFFFF}\begin{tabular}[c]{@{}c@{}}\textbf{Height}\\ (meters)\end{tabular}} \\ \cline{2-7} 
\multirow{-2}{*}{} &
  \multicolumn{1}{c|}{\cellcolor[HTML]{EFEFEF}-4.5} &
  \multicolumn{1}{c|}{\cellcolor[HTML]{EFEFEF}-5} &
  \cellcolor[HTML]{EFEFEF}-5.5 &
  \multicolumn{1}{c|}{\cellcolor[HTML]{EFEFEF}1.585} &
  \multicolumn{1}{c|}{\cellcolor[HTML]{EFEFEF}1.635} &
  \cellcolor[HTML]{EFEFEF}1.685 \\ \hline \hline
\rowcolor[HTML]{FFFFFF} 
\multicolumn{1}{|c|}{\cellcolor[HTML]{FFFFFF}\begin{tabular}[c]{@{}c@{}}\textbf{Errors}\\ (meters)\end{tabular}} &
  \multicolumn{1}{c|}{\cellcolor[HTML]{FFFFFF}4.05} &
  \multicolumn{1}{c|}{\cellcolor[HTML]{FFFFFF}0} &
  2.71 &
  \multicolumn{1}{c|}{\cellcolor[HTML]{FFFFFF}0.73} &
  \multicolumn{1}{c|}{\cellcolor[HTML]{FFFFFF}0} &
  0.72 \\ \hline
\end{tabular}%
}
\end{table}

\begin{figure}[htbp]
	\centering        
        \includegraphics[width=0.47\textwidth]{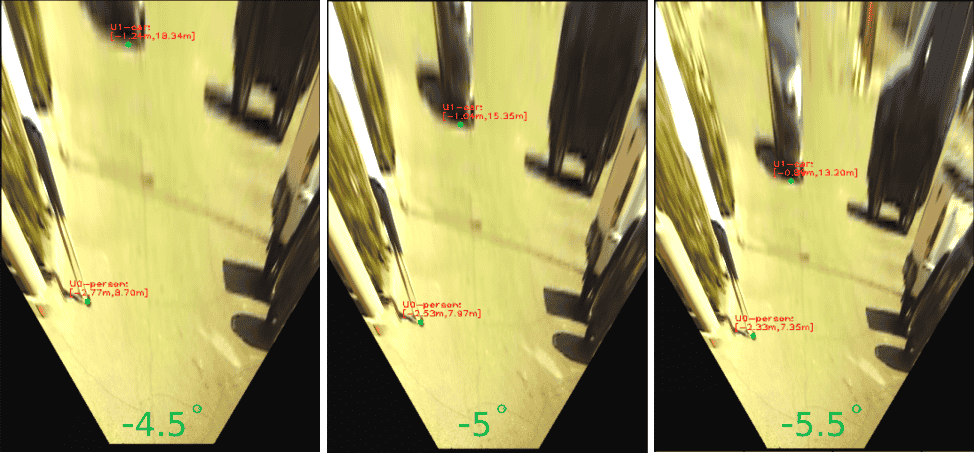}
	\caption{Visualization of Errors Caused by Pitch Angle Change. } 
	\label{bev}
\end{figure}
\begin{figure}[htbp]
	\centering        
        \includegraphics[width=0.47\textwidth]{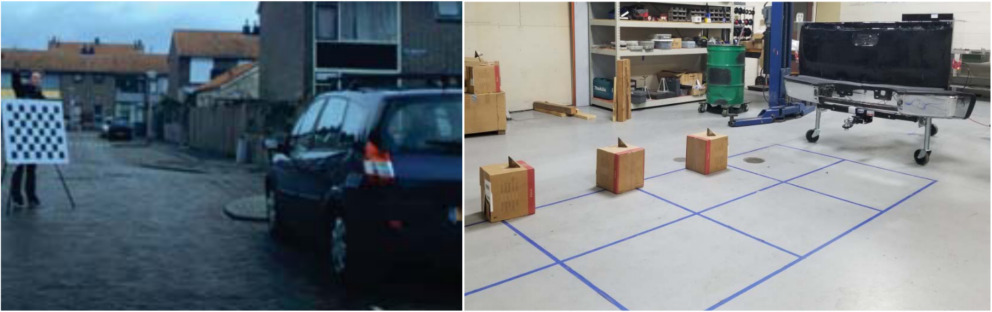}
	\caption{Target-Based Calibration Scenarios from \cite{domhof2019extrinsic} and \cite{olutomilayo2021extrinsic} typically require specific targets and environments, as well as manual intervention. } 
	\label{target_cali}
\end{figure}

Despite the advantages of sensor fusion, traditional radar-camera fused MOT grapples with several limitations. First, radar point cloud data is typically too sparse \cite{wang2021rodnet} to extract meaningful features. Additionally, the clustering methods used for generating radar object detection, such as Density-Based Spatial Clustering of Applications with Noise (DBSCAN), require manual fine-tuning for optimal performance \cite{palffy2020cnn}. This process can be time-consuming and may not always produce reliable results. Second, the camera branch often requires the implementation of Bird's Eye View (BEV) transformation to infer the real-world positions of objects. 
This transformation necessitates manual measurements of the camera's height and pitch angle, which can be error-prone \cite{dimitrievski2019people}. In fact, the monocular camera's BEV transformation is highly sensitive to measurement errors, as shown in Table \ref{tabbev}. Even slight differences in height or pitch angle can result in significant errors (see Fig. \ref{bev}) in distance estimation. Third, the association between radar and camera data requires precise calibration to ensure accurate alignment, which is usually achieved through target-based methods that rely on manual intervention \cite{cheng2023online}, as shown in Fig.  \ref{target_cali}. For example, some methods require measuring the installation position and angle between the radar and camera \cite{bai2021robust}, while others involve manually adjusting the calibration target's position \cite{domhof2019extrinsic,olutomilayo2021extrinsic,cheng20233d}.

To overcome these shortcomings, we propose an efficient radar-camera fusion MOT framework that minimizes manual interventions. Herein, the radar functions as the primary sensor to provide precise range measurements, while the camera complements its functions by providing object classification information. We select the low-level Range-Azimuth-Doppler (RAD) tensor data from radar, which contains rich information about radar detections \cite{wang2021rethinking,wang2021rodnet} and resembles image data, to fully exploit radar data and identify common features with camera data. Building on our prior work \cite{cheng2023online}, these common features enable us to implement an online targetless calibration method, which associates radar and camera data without manual intervention. This online calibration allows for the straightforward projection of camera detections onto the radar Range-Azimuth (RA) plane, directly ascertaining the real-world positions of camera-detected targets within the radar's field of view, thereby bypassing the need for intricate camera BEV transformations. Additionally, leveraging common features allows for not just position but also feature matching between radar and camera, enhancing the precision of the association outcomes. The contributions of this work are highlighted as follows: 

\begin{enumerate}
  \item  We propose an efficient radar-camera fusion MOT framework that minimizes manual intervention by eliminating the need to manually measure sensor installation positions and angles or to move calibration targets, thus streamlining the fusion of radar and camera data.
  \item  This method exploits the common features between low-level radar RAD data and camera data to establish an online targetless calibration between the two sensors, enabling the direct derivation of real-world positions of camera-detected objects without relying on error-prone camera BEV transformations. % within the radar's field of view
  \item  By combining position matching and common features-based feature matching, we achieve a more robust fusion of radar and camera detections, surpassing the limitations of mere position matching. Furthermore, we employ category-consistency checking to associate objects across consecutive frames, enhancing inter-frame object association accuracy.
\end{enumerate}

The rest of the paper is organized as follows: Section II provides background information. Section III reviews the related works. Section IV details our proposed MOT approach. Section V presents our experimental evaluations. Section VI concludes the paper.
\begin{figure*}[h!]
	\centering        
        \includegraphics[width=0.98\textwidth]{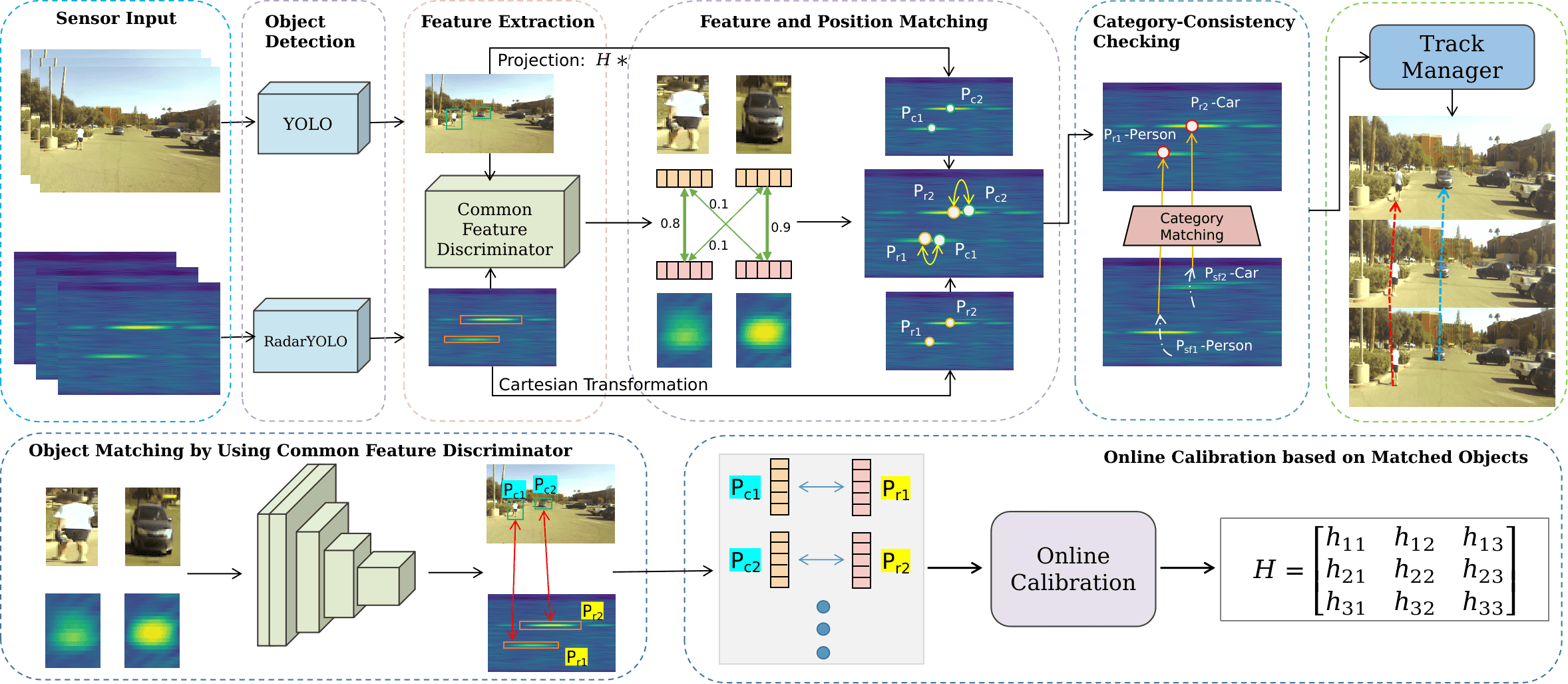}
	\caption{Framework of the Proposed Radar-Camera Fusion MOT Method.}
	\label{0}
\end{figure*}

\section{Background}
\subsection{Traditional Radar-Camera Fusion-Based MOT}
The core aim of MOT is to accurately estimate the trajectories of multiple objects and re-identify them across consecutive frames captured by one or multiple sensors \cite{luo2021multiple}. Achieving this goal requires the development of algorithms capable of addressing three key challenges: locating multiple objects, maintaining their identities (IDs), and determining their individual trajectories \cite{luo2021multiple}. With recent advancements in object detection, the tracking-by-detection paradigm has been the dominant approach in MOT\cite{meinhardt2022trackformer}. This approach involves firstly using the object detectors to extract detections from both sensors, and then associating them across frames to assign unique IDs to each object \cite{ciaparrone2020deep}. In the context of radar-camera fused MOT, the approach typically comprises three main branches:

\subsubsection{Radar Branch}
In the radar branch, radar point clouds are often utilized as input data, typically necessitating the use of clustering algorithms to obtain clusters corresponding to different targets, i.e., radar detections. The DBSCAN algorithm, known for its ability to handle outliers and noisy datasets without requiring the number of clusters to be specified \cite{deng2022robust}, is widely adopted in this context. Subsequently, motion estimation algorithms like the Kalman filter are employed for trajectory prediction, followed by the utilization of the Hungarian algorithm to associate detections in the next frame with existing trajectories \cite{wang2023camo}. The performance of the radar branch largely hinges on the effectiveness of clustering. However, clustering is error-prone, often leading to incorrect merging or splitting of objects \cite{palffy2020cnn}. Determining appropriate parameters, such as radius and minimum point count for DBSCAN, is a difficult task since the same parameters must be applied to all classes despite their vastly different spatial extents and velocity profiles \cite{palffy2020cnn}. Moreover, clustering methods often struggle to extract useful features, such as variance, from small objects that lack sufficient reflections \cite{palffy2020cnn}. Following the clustering process, the mean position of the cluster (i.e., the cluster centroids) is calculated, representing the position of the detection. 

\subsubsection{Camera Branch}
In the camera branch, the process begins with object detection using image detectors, which typically output bounding boxes. The lower centers of these bounding boxes are used as the objects' positions. However, unlike radar, the camera doesn't directly provide the real-world spatial location of objects and requires additional transformations. To determine an object's real-world position from image detections, a common approach involves converting from the image plane to the 2D ground plane, or BEV, using inverse perspective mapping \cite{jeong2016adaptive}. This method, while prevalent, often requires some prior information such as camera height, pitch angle, and average human height \cite{bai2021robust,cui2023online,sengupta2022robust,dimitrievski2019people}, and thus involves manual measurement. Once the object’s position on the 2D ground plane is ascertained, the camera branch then proceeds to motion estimation and track-detection association, similar to those in the radar branch.
\subsubsection{Sensor Fusion Branch}
In this branch, a decision-level fusion is typically employed to combine the real-world positions of objects detected in the radar and camera fields of view. This integration requires calibration between the two modalities to ensure accurate matching of objects detected by both sensors. By fusing the outputs of the two sensors, the MOT system can benefit from the strengths of both sensors, leading to improved robustness and performance. Specifically, when one sensor fails or experiences a detection gap, the other sensor's detection can supplement the missing information, enhancing the overall perception capability of the system. The success of this fusion heavily relies on the quality of the association between radar and camera detections, with effective radar-camera calibration being key to reliably associating objects identified by both sensors. This calibration—is typically achieved using a target-based approach—demands manual intervention, making it laborious and impractical.

\subsection{Radar-Camera Calibration}
Radar-camera calibration enables the association of the same target detected by these two sensors in their respective coordinate systems. This process involves solving a calibration matrix that can transform a point in the 3D radar coordinate system (RCS) to the 2D image pixel coordinate system (PCS) \cite{cheng20233d,cheng2023online}. Its homogeneous transformation can be represented as:
\begin{equation}\label{eq3}
%\small
\begin{split}
s
\begin{bmatrix}
  u_r  \\
  v_r  \\
  1
\end{bmatrix} 
&= K_{3 \times 3}
\begin{bmatrix}
  R_{3 \times 3} | T_{3 \times 1} \\
\end{bmatrix}
\begin{bmatrix}
  x_r  \\
  y_r  \\
  z_r  \\
  1
\end{bmatrix}
\\
&=
\begin{bmatrix}
  f_x & \gamma & u_0 \\
  0 & f_y & v_0 \\
  0 & 0 & 1 \\
\end{bmatrix}
\begin{bmatrix}
  r_{11} & r_{12} & r_{13} & t_x \\
  r_{21} & r_{22} & r_{23} & t_y \\
  r_{31} & r_{32} & r_{33} & t_z \\
\end{bmatrix}
\begin{bmatrix}
  x_r  \\
  y_r  \\
  z_r  \\
  1
\end{bmatrix}, 
\end{split}
\end{equation}
where  $P_p=(u,v)$ is a point in PCS, $P_r=(x_r,y_r,z_r)$ is a point in RCS, $\hat{P}_r=(u_r,v_r)$ is the $P_r$'s 2D projection in PCS, $f_x$ and $f_y$ are the camera focal lengths, $(u_0,v_0)$ represent the principal point, $s$ is the scaling factor, $\gamma$ is the skew coefficient between the image axes \cite{szeliski2022computer}, $K_{3 \times 3}$ is the camera intrinsic matrix, $R_{3 \times 3}$ and $T_{3 \times 1}$ are the rotation and translation matrix.

To determine this calibration matrix, we need to collect a set of \(N\) points in the RCS and their corresponding points in the PCS. Although 6 points are sufficient to solve the matrix \cite{hartley2003multiple}, we typically use more points to allow for more adjustments and estimations, and to account for measurement errors, thereby increasing precision. The optimal solution is determined by minimizing the geometric reprojection error \cite{hartley2003multiple,dubrofsky2009homography}, which is also known as solving the Perspective-n-Point (PnP) \cite{marchand2015pose} problem. The reprojection error can be considered the root-mean-square residual error between $N$ of the projected radar 2D pixel points and the corresponding image 2D pixel points, and can be defined as:
\begin{equation}\label{eq9}
\begin{split}
\mathcal{E}_{rep}
&=\sqrt{\frac{1}{N}\sum_{i=1}^{N}\left\| P^{i}_{p}-\hat{P}^{i}_{r} \right\|_{2}^{2}}\\
&=\sqrt{\frac{1}{N}\sum_{i=1}^{N}(u^{i}-u^{i}_r )^{2} + (v^{i}-v^{i}_r )^{2}}.
\end{split}
\end{equation}

Target-based and targetless methods are commonly used to collect radar-camera point pairs for calibration. Target-based methods use calibration targets, such as corner reflectors, that can be clearly detected by both radar and camera, to collect point pairs. However, these methods necessitate the preparation of specific calibration scenes and manual adjustment of the calibration target positions, as shown in Fig. \ref{target_cali}. On the other hand, targetless methods do not rely on calibration targets but instead achieve point pair collection by matching natural features present in the scene that are observable by both the radar and camera \cite{scholler2019targetless}. Our work adopts the targetless method, leveraging its convenience, flexibility, and ability to operate without manual intervention.

%\subsection{Exploring Common Features in Radar and Camera through Deep Learning}
\subsection{Deep Learning For Radar Data}
Targetless calibration reduces manual effort and controlled setups but still relies heavily on accurately extracting high‑quality natural features\cite{cheng2023online}, a task made especially difficult by low‑resolution, high‑noise radar data.
However, recent advances in deep learning have made radar feature extraction increasingly promising \cite{bhattacharya2020deep}. Radar data typically takes two forms—point clouds and RAD cubes \cite{harlow2023new}. Radar point clouds hold limited information, making it difficult for deep learning models to derive semantic meaning. In contrast, dense RAD data—rich in Doppler information and surface textures \cite{wang2021rodnet}—offers stronger semantic understanding of radar detections, and its image‑like format further allows us to exploit mature computer‑vision deep‑learning techniques; so we chose to use RAD data in this study.

Indeed, recent works have demonstrated the success of deep learning on RAD data across classification, detection, and tracking tasks. \cite{bhattacharya2020deep} showed that CNNs can effectively classify radar objects using RAD cubes. Subsequent work—including RODNet\cite{wang2021rodnet}, RADDet\cite{zhang2021raddet}, and TransRAD\cite{TransRAD}—has proven that these models can learn from dense RAD data to achieve accurate object detection. A YOLOv4\cite{bochkovskiy2020yolov4}-based architecture was also adapted for object tracking on Range-Azimuth (RA) maps \cite{kim2022deep}. Moreover, our previous work \cite{cheng2023online} demonstrated that a YOLO-based model could effectively extract useful features from RAD data and perform common feature extraction. Building on this foundation,  our current work aims to enhance interoperability and cooperation between radar and camera for MOT by leveraging deep learning's ability to uncover shared representations across these two sensors.

\begin{figure*}[h!]
	\centering        
        \includegraphics[width=0.86\textwidth]{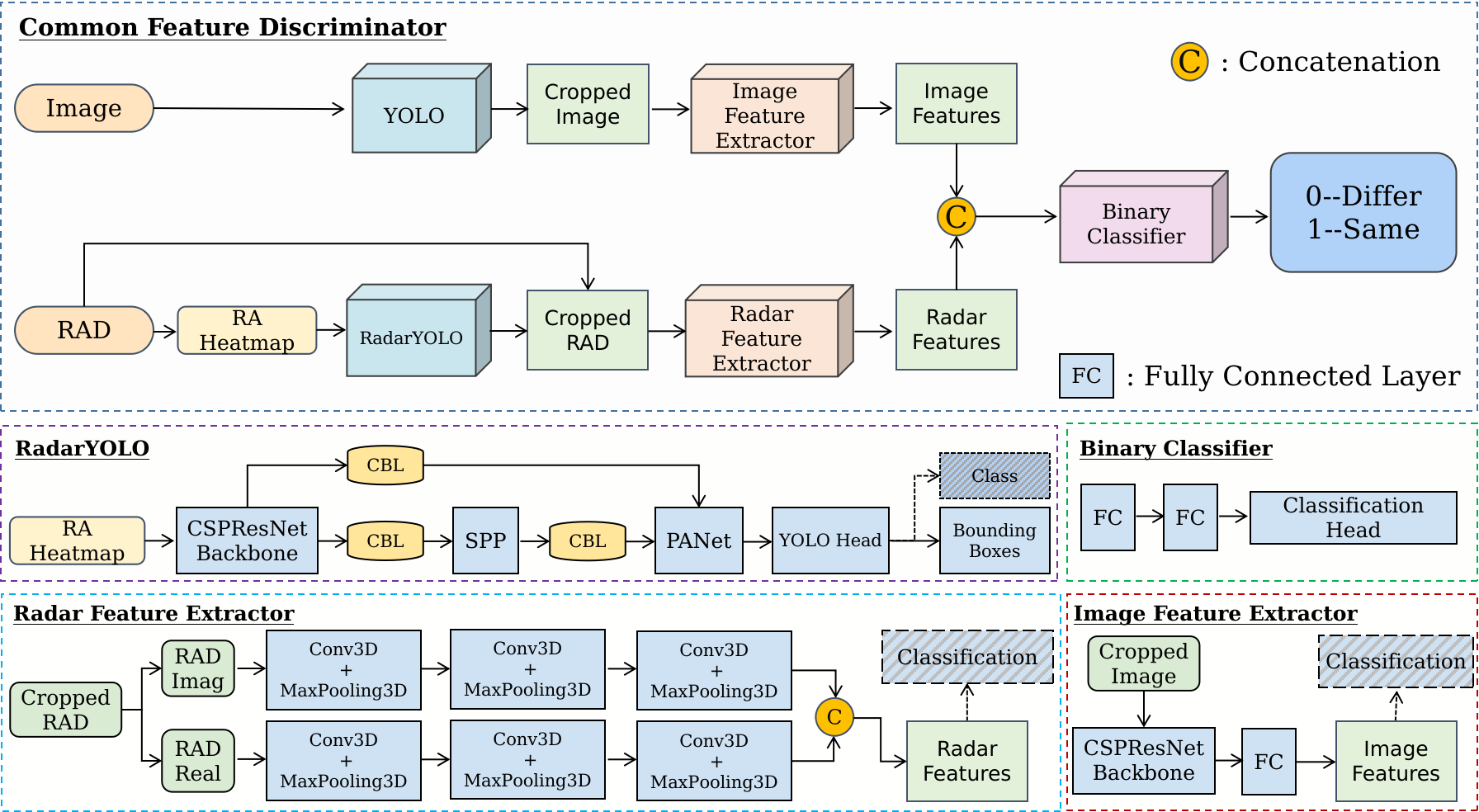}
	\caption{Common Feature Discriminator Model Architecture. CSPResNet: Cross-Stage Partial ResNet, CBL: Convolution3D + Batch Normalization + LeakyReLU, SPP: Spatial Pyramid Pooling, PANet: Path Aggregation Network, are from YOLO \cite{bochkovskiy2020yolov4}.}
	\label{com}
\end{figure*}

\section{Related Works} % And Motivations
\label{related_work}
Currently, most MOT methods remain camera-centric, tracing back to the SORT algorithm \cite{bewley2016simple}, which established the tracking-by-detection paradigm via a CNN detector, Kalman filter motion estimator, and Hungarian matcher. Subsequent work has validated or refined its components: a basic Kalman filter alone can achieve state-of-the-art motion estimation \cite{cao2023observation}; appearance information, emphasized in Deep SORT \cite{wojke2017simple}, greatly reduces identity switches; hierarchical association schemes further improve data matching \cite{zhang2023bytetrackv22d3dmultiobject}; and strong detectors boost tracking-by-detection MOT performance \cite{zhang2023motrv2}. Inspired by these successful strategies, our framework also adopts the tracking-by-detection paradigm. Specifically, we employ a YOLOv4 for camera and RadarYOLO for radar, use a Kalman filter for motion estimation, and apply the Hungarian algorithm for data association. Additionally, we explicitly leverage distinctive appearance features across radar and camera modalities to facilitate more accurate radar-camera and detection-track associations. 

Despite achieving promising results, camera-only MOT methods falter in poor lighting or adverse weather, spurring efforts to fuse complementary sensors to enhance the robustness and reliability of MOT systems. One such direction involves fusing thermal infrared (TIR) sensors with cameras. TIR is one such modality—its lighting-invariant imagery \cite{zhang2024comprehensive} pairs well with RGB data and has been fused via self-supervised transformers \cite{liu2024stfnet} or hierarchical attention networks \cite{yuan2024hierarchical} to improve tracking in low-light scenes.
However, a notable limitation of thermal infrared sensors is their inability to provide distance estimation, which is crucial for applications in traffic monitoring and autonomous driving. Unlike TIR sensors, ranging sensors such as radar and LiDAR are capable of accurate range measurements. Nevertheless, while TIR images are visually similar to RGB images and can be fused efficiently, the data characteristics of ranging sensors differ significantly from camera data, making effective fusion of them for MOT a persistent research challenge.

Currently, many state-of-the-art MOT systems based on distance sensing rely on LiDAR-camera fusion \cite{wang2023camo}. However, LiDAR has several inherent drawbacks compared to radar, including higher cost, limited detection range \cite{li2020lidar}, and, most importantly, the inability to measure object velocity—an essential factor for robust tracking. These limitations make radar-camera-based MOT systems increasingly attractive, and they form the core focus of this work.

Recent studies have explored radar–camera fusion to boost MOT. Authors in \cite{bai2021robust} perform independent radar and camera detection, then correlate the detection results within the image plane for tracking. Authors in \cite{liu2021robust} fuse radar position with camera classification via Mahalanobis distance and a joint-probability function. Authors in \cite{deng2022robust} use vision back-projection to let radar and camera serve as mutual references to deduce object positions. Authors in \cite{sengupta2022robust} combine high-level camera and radar outputs with decision-level fusion and a tri-Kalman filter for resilience to single-sensor failures. Authors in \cite{leimot} leveraged a Bi-LSTM network for improved motion prediction and a FaceNet-inspired appearance model to enhance detection-track associations, employing a tri-output mechanism to provide redundancy against sensor failures.

Despite their advances, these methods underuse radar data—either discarding low-level information via traditional detection pipelines that output only final target lists, or depending on DBSCAN clustering of sparse point clouds, which demands manual parameter tuning and can be unreliable. Some recent efforts have tapped richer RAD data for pedestrian or vehicle tracking \cite{dimitrievski2019people,ram2022fusion}, but these tend to focus on specific object classes and diverge from the mainstream MOT framework, and few address sensor failures explicitly.

Critically, many of these radar-camera methods overlook detailed descriptions of calibration procedures or how to obtain accurate real-world positions of camera-detected objects. When calibration methods are discussed, they commonly involve manual interventions, making them impractical for real-world applications. For example, BEV-based methods (\cite{bai2021robust, sengupta2022robust, cui2023online, wang2021rethinking}) require manual measurements of camera height or target dimensions; the approaches in \cite{dimitrievski2019people, kim2021robust} require measuring average person heights or relative scene geometry; and others \cite{ram2022fusion, deng2022robust} require manual checkerboard calibration, respectively. All these approaches typically rely on target-based calibration setups, which necessitate controlled lab conditions and significant manual intervention, severely limiting their practical applicability.

In contrast, our proposed framework is explicitly designed to overcome these critical limitations. Distinctively, we position radar as the primary sensor for object localization by developing a specialized deep-learning-based radar detector—RadarYOLO—which directly processes low-level radar RAD data to generate accurate object positions. This effectively avoids the shortcomings associated with sparse radar point clouds and the manual tuning required by clustering methods. Furthermore, we introduce a novel online targetless calibration approach driven by deep-learning-extracted common features shared between radar RAD data and camera images. To our knowledge, this represents the first exploitation of radar-camera common features for automatic, runtime calibration within an MOT system. Our method seamlessly establishes reliable radar-camera correspondences without manual interventions, significantly simplifying calibration and reducing camera object position-deducing errors associated with traditional BEV transformations. Lastly, we propose a robust multi-stage matching strategy in the sensor fusion branch, combining feature-based similarity assessment, spatial proximity matching, and a category-consistency check. This strategy significantly surpasses traditional position-only association methods, providing greater tracking accuracy, improved inter-frame association reliability, and enhanced robustness against single-sensor failures.

\begin{figure*}[htbp]
	\centering        
        \includegraphics[width=0.999\textwidth]{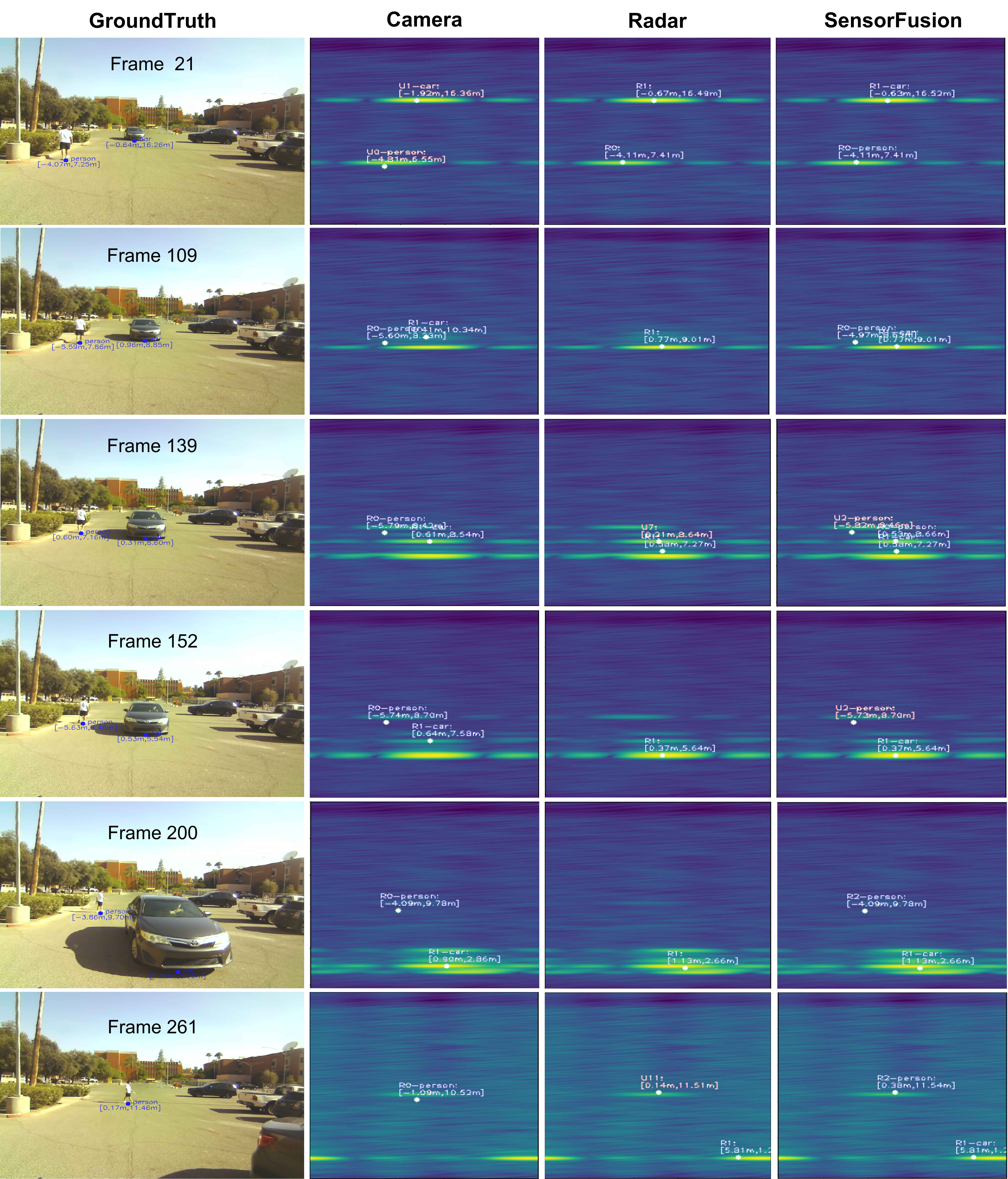}
	\caption{Example Frames Demonstrating Multi-Object Tracking Results of the Proposed Method. Camera Branch Results are Obtained by Projecting Camera Detections onto the Radar Plane.} 
	\label{4}
\end{figure*}

\section{Proposed Method}
The proposed radar-camera fusion MOT framework, schematically illustrated in Fig. \ref{0}, adheres to the classic "tracking-by-detection" paradigm and encompasses three integral branches—the radar branch, the camera branch, and the sensor fusion branch—each performing dedicated yet complementary roles to enhance tracking robustness and mitigate single-sensor failures. In the radar branch, our specially developed RadarYOLO detector processes radar data represented as RA heatmaps and outputs object bounding boxes in the radar RA plane. The centers of these bounding boxes provide precise spatial positions of objects in polar coordinates, which are subsequently transformed into Cartesian coordinates to directly obtain physically meaningful object locations. For the camera branch, the YOLO detector \cite{bochkovskiy2020yolov4} identifies and localizes objects within image frames. However, as camera detections inherently reside in the two-dimensional image plane without direct real-world spatial correspondence, it is necessary to convert these detections into physically interpretable coordinates. To achieve this, we implement an online targetless radar-camera calibration method based on learned common features extracted from radar and camera data (as detailed previously). This calibration enables a direct and automatic mapping of camera-detected image points onto the radar RA plane to derive their real-world positions. In the sensor fusion branch, decision-level fusion is utilized to integrate radar and camera detections, effectively combining the distinct advantages offered by each sensor modality—namely, radar's accurate distance measurements and camera's robust object classification capabilities. This process includes a two-stage matching approach, comprising common feature-based similarity assessments followed by spatial proximity matching, to robustly associate radar and camera detections. After association, sensor fusion detections inherit spatial locations from the radar branch, leveraging radar’s superior spatial accuracy, while adopting object categories from the camera branch. Additionally, a category-consistency check strategy is incorporated into the detection-track association, effectively preventing tracking errors that commonly arise in traditional position-only association methods, thus significantly enhancing inter-frame tracking stability and accuracy. Finally, a Kalman filter-based motion estimation module predicts the future positions of tracked objects, complemented by a proven track management approach previously introduced in our earlier work \cite{leimot}, to further ensure continuity and stability in multi-object tracking outcomes.

\subsection{Radar-Camera Common Features}
Radar and camera sensors, due to their distinct physical working principles, produce sensing data with different characteristics. Radar RAD data typically provides unique features that can identify objects, such as reflectivity intensity, frequency distribution, Doppler motion characteristics, and derived distance, velocity, and angular features. In contrast, camera image data captures object-specific features like color, texture, edges, contours, and spatial correlations. Although these two types of sensing data seem unrelated, they share some commonalities: 
\begin{enumerate}
  \item  Reflection Intensity: Like radar, cameras also collect information on the reflection intensity of an object.
  \item  Shape and Texture Recognition: Radar reflection intensity, frequency information, and range-angle spectra can help identify the material and shape of objects, indirectly inferring texture and contours.
  \item  Semantic Information: Radar data can reveal high-level semantic information that matches with image data. For example, a car detected in the image might correspond to an object with prominent Doppler features in the radar.
  \item  Statistical Properties: The statistical properties of radar and camera data may exhibit similarities.
\end{enumerate}
Thus, discovering and utilizing these shared characteristics observed in radar and camera detections of the same objects is possible. To achieve this, we leverage deep learning's superior feature extraction capabilities to develop a model called the Common Feature Discriminator, designed to learn and extract shared features between radar and camera data.

The Common Feature Discriminator, as shown in Fig. \ref{com}, takes two inputs: the radar RAD data of an object and the image data of either the same or a different object. The radar object RAD data is processed by a radar feature extractor, which uses a 3D convolutional neural network and convolves the real and imaginary parts of the radar data separately to extract 128-dimensional radar embedding features. A classification head is employed during training to aid convergence but is disabled when integrated with the Common Feature Discriminator. Similarly, the image object is processed by an image feature extractor, which consists of the backbone of YOLOv4, a fully connected layer, and a classification head (disabled during integration), to obtain a 128-dimensional image embedding feature. The radar and image embedding features are then concatenated and input into a binary classifier to determine whether the two objects are the same.

\begin{figure}[htbp]
	\centering        
        \includegraphics[width=0.498\textwidth]{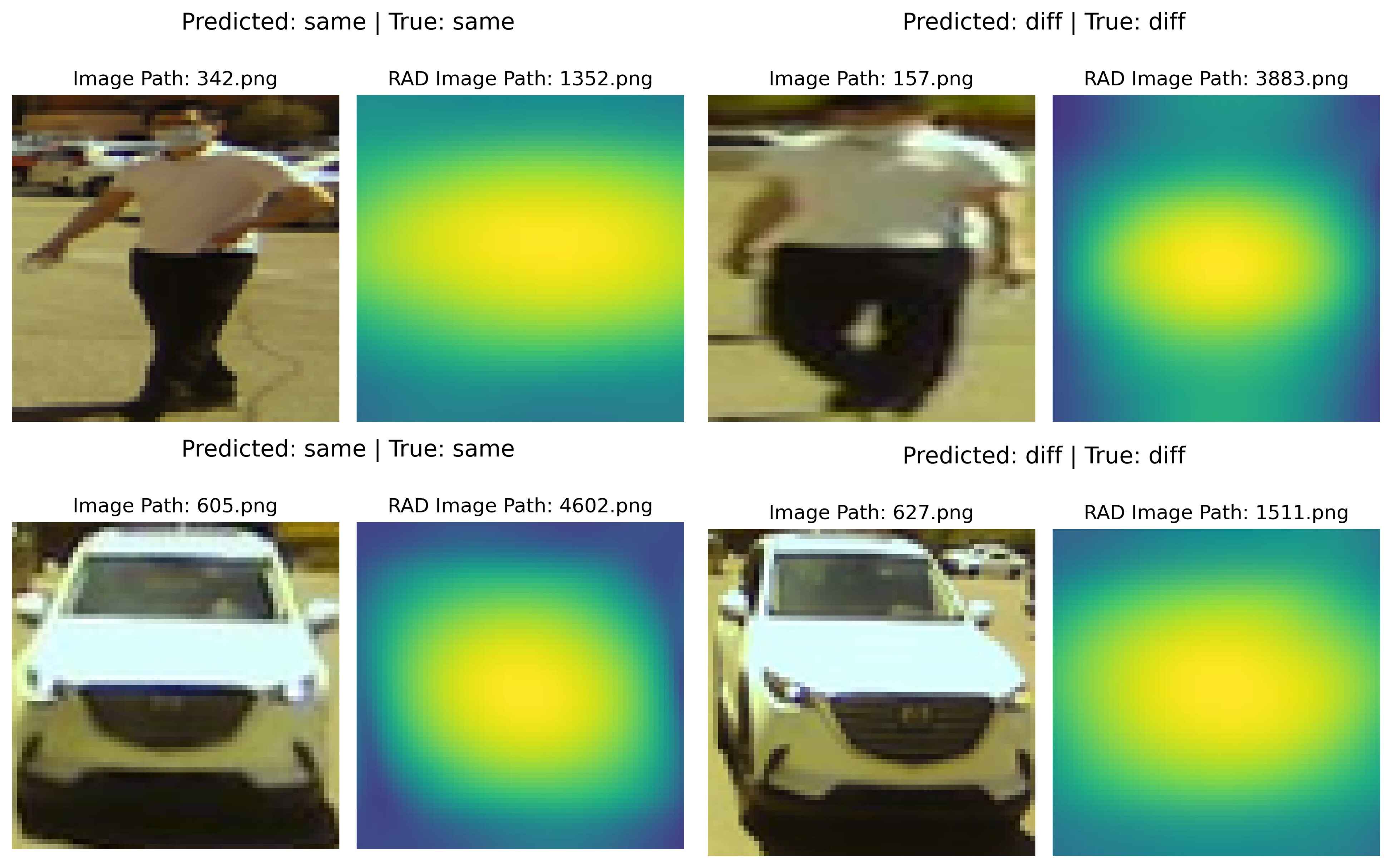}
	\caption{Test Results using the Common Feature Discriminator.} 
	\label{com_feat_res}
\end{figure}

To enable end-to-end inference, where the input is the entire sensor frame data (rather than cropped object data) and the output is the judgment of whether two objects are the same, the radar detector and image detector are developed to work in conjunction with the Common Feature Discriminator, forming a complete end-to-end object matching system.
Specifically, time-aligned radar and image frames are processed by the radar and image detectors to identify all objects in their frames. The image detector uses the official pretrained YOLOv4 model directly due to its already excellent performance on our data, to leverage its proven feature extraction and object detection capabilities, while the radar detector, named RadarYOLO, is adapted from YOLOv4 and uses RA maps as input. These detectors perform object detection on their respective sensor frames and output the detected objects' 2D bounding boxes. Based on these bounding boxes, the sensor data corresponding to each object is cropped. The cropped radar data includes the entire Doppler dimension, assuming that there are no multiple objects with different velocities within the same RA bounding box. The cropped single-object radar and camera data are then fed into the Common Feature Discriminator to match the same objects and reject different ones, as shown in Fig. \ref{com_feat_res}.

Note that RadarYOLO was initially pretrained on the publicly available RADDet dataset \cite{zhang2021raddet}, which consists of 10,158 frames and 28,401 objects from six classes: person, bicycle, car, motorcycle, bus, and truck, with an 80/20 training-test split. To address variations arising from different radar configurations, we manually annotated 2,200 frames of our collected radar data (including 6,335 radar objects classified as person, car, and truck) for transfer learning. Simultaneously, corresponding camera objects were annotated to train the Common Feature Discriminator. The discriminator was trained after freezing the pretrained YOLOv4 and RadarYOLO weights. Training was performed using TensorFlow on an Nvidia 32GB V100S GPU with an AMD Zen2 processor. RadarYOLO was trained for 300 epochs using SGD with momentum 0.937, batch size 4, and a cosine learning rate (initial: 0.01, minimum: 0.0001). The Common Feature Discriminator was trained for 800 epochs using Adam optimizer with momentum 0.9, weight decay 0.0004, batch size 4, and a cosine learning rate (initial: 0.001, minimum: 0.00005).

\renewcommand{\algorithmiccomment}[1]{\hfill$\triangleright$\textit{\textcolor{blue}{#1}}}
\begin{figure}[htbp]
    \centering
    \begin{minipage}{0.498\textwidth}
        \begin{algorithm}[H]
            \caption{Online Calibration Based on Common Features}\label{alg:onlinecalib}
            \textbf{Input: } $\text{Frames} = \{(R_i, C_i)\}_{i=1}^{T}$ \Comment{$T$ frames of sensor data}\\
            \quad\quad\;\; $\text{BS}$ \Comment{Sampling block size (pixels)}\\
            \quad\quad\;\; $\text{H}_{img}$ \Comment{Height of camera image}\\
            \quad\quad\;\; $\bm{\mathcal{M}}$ \Comment{Common Feature Discriminator-based matcher}\\
            \textbf{Output: } $\mathbf{H}_{upper}, \mathbf{H}_{lower}$ \Comment{Two calibration matrices}

            \begin{algorithmic}[1]
                \State $\text{PairsUpper},\, \text{PairsLower} \gets [\,], [\,]$ \Comment{Store correspondences}

                \For{$i \in \{1,2,\dots,T\}$}
                    \State $\text{Objs}_r \gets \text{RadarYOLO}(R_i)$ \Comment{Radar objects}
                    \State $\text{Objs}_c \gets \text{YOLOv4}(C_i)$ \Comment{Camera objects}

                    \State $\text{MatchPairs} \gets [\,]$ \Comment{Corresponding points}
                    \For{each $(O_r, O_c) \in \text{Objs}_r\times \text{Objs}_c$}
                        \If{$\bm{\mathcal{M}}(O_r, O_c)$ is True}
                            \State $\text{MatchPairs.append}(O_r^{RA}, O_c^{img})$ 
                        \EndIf
                    \EndFor
                    
                    \State $\text{SampledPairs}\gets\text{BlockBasedSampling }(\text{MatchPairs}, \text{BS})$
                    \For{each $(O_r^{RA}, O_c^{img}) \in \text{SampledPairs}$}
                        \If{$O_c^{img}(v) < \frac{2}{3}\text{H}_{img}$} \Comment{Upper (distal) points}
                            \State $\text{PairsUpper.append}(O_r^{RA}, O_c^{img})$ 
                        \Else \Comment{Lower (proximal) points}
                            \State $\text{PairsLower.append}(O_r^{RA}, O_c^{img})$ 
                        \EndIf
                    \EndFor
                \EndFor
                \State
                $\mathbf{H}_{upper}\gets\text{EstimateHomography}(\text{PairsUpper})$
                \Comment{Solve homography with RANSAC}
                \State $\mathbf{H}_{lower}\gets\text{EstimateHomography}(\text{PairsLower})$
                \Comment{Solve homography with RANSAC}

                \State\Return $\mathbf{H}_{upper}, \mathbf{H}_{lower}$

            \end{algorithmic}
        \end{algorithm}
    \end{minipage}
\end{figure}

\subsection{Online Calibration Based On Common Features}
\begin{figure}[htbp]
	\centering        
        \includegraphics[width=0.499\textwidth]{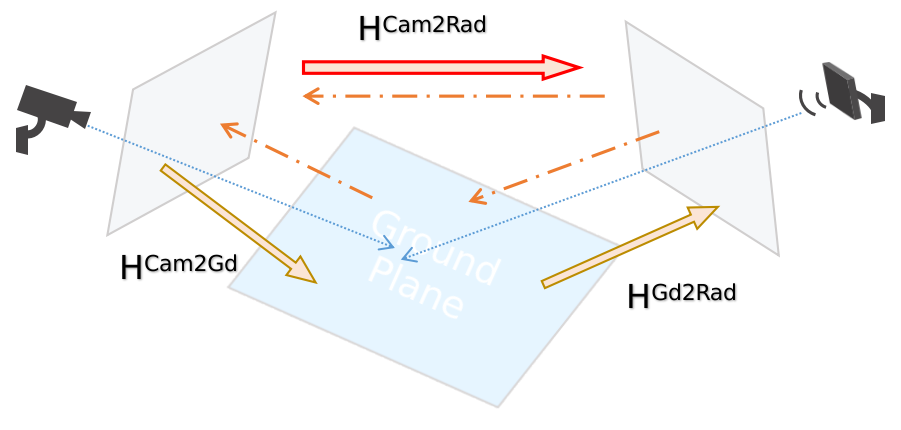}
	\caption{Homography Transformation between Radar and Camera Planes.} 
	\label{homog}
\end{figure}

Most existing radar-camera calibration methods utilize a complex 3D-to-2D perspective projection. In contrast, we propose using the planar projective transformation to achieve 2D-to-2D calibration between the radar RA plane and the camera image plane. A planar projective transformation, or Homography, is an invertible linear transformation represented by a non-singular matrix  $\mathbf{H} \in \mathbb{R}^{3 \times 3}$ \cite{dubrofsky2009homography,nguyen2018unsupervised}. With this matrix, we can project a point in the PCS onto the radar RA plane as follows:
\begin{equation}
\label{eq_H}
\begin{bmatrix}
r_p \\
\theta_p \\
1 
\end{bmatrix} 
= \mathbf{H}
\begin{bmatrix}
u \\
v \\
1 
\end{bmatrix}
= 
\begin{bmatrix}
h_{11} & h_{12} & h_{13} \\
h_{21} & h_{22} & h_{23} \\
h_{31} & h_{32} & h_{33}
\end{bmatrix}
\begin{bmatrix}
u \\
v \\
1 
\end{bmatrix},
\end{equation}
where \( \hat{P_p} = (r_p, \theta_p) \) is the projection point of \( P_p \) in the radar RA polar coordinate system.
Notably, the objects or points on the radar RA plane, as well as those on the camera image plane, are derived from objects or points all lying on a common plane (i.e., the ground plane), as illustrated in Fig. \ref{homog}. Therefore, this radar-camera calibration can be considered as the planar homography induced by a common plane \cite{szeliski2022computer}, justifying the use of 2D Homography for it.

Typically, to estimate this homography calibration matrix, we use point correspondences. By leveraging the end-to-end Common Feature Discriminator-based object matching system \( \bm{\mathcal{M}}(f_r, f_c) \), where \( f_r \) and \( f_c \) represent feature vectors from radar and camera detections respectively, we can perform object matching and correspondence identification without using any calibration targets (targetless). By comparing the features of objects detected in radar and camera views, we can identify corresponding objects across modalities:
\begin{equation}
f_r \overset{\bm{\mathcal{M}}}{\sim} f_c \Rightarrow P_r^{RA} \leftrightarrow P_p,
\end{equation}
where \( P_r^{RA} = (r, \theta) \) is the position of an object in the radar RA polar coordinate system. Based on these point correspondences, the Direct Linear Transform \cite{hartley2003multiple} is used to solve for the calibration matrix, with RANSAC employed to exclude outliers and achieve robust results \cite{dubrofsky2009homography}.

\begin{figure*}[]
	\centering        
        \includegraphics[width=0.93\textwidth]{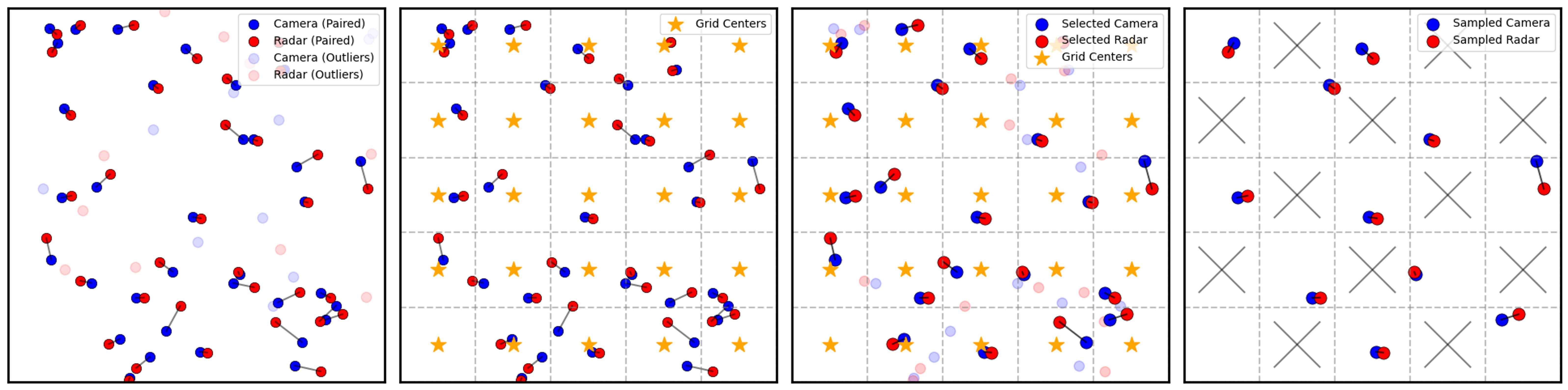}
	\caption{Block-Based Sampling Strategy: (1) Project radar points onto the image to form radar-camera point pairs; (2) Divide image into equal-sized grids; (3) Retain point pairs closest to each grid center; (4) Sample pairs at intervals of one block.} 
	\label{blk}
\end{figure*}

Furthermore, to ensure that the point correspondences used to solve the calibration matrix are well-distributed across the sensor field of view, making the calibration results more representative and generalizable \cite{cheng2025calibrefine}, we employed a block-based sampling strategy, as illustrated in Fig. \ref{blk}. This strategy partitions the camera image plane into many small blocks, each measuring $5 \times 5$ pixels. The image-radar point correspondences collected based on common feature matching are assigned to these blocks based on their spatial location within the image. Specifically, we selected one block at intervals of one block and identified the point correspondence closest to the center of that block as the representative point correspondence for that block, discarding the other points. This procedure yielded a set of representative point correspondences that captured the spatial distribution, enabling us to obtain accurate calibration results.
Notably, the proposed calibration method requires no manual intervention, allowing for online real-time calibration, and thus can handle runtime decalibration. For example, during sensor operation, point correspondences collected over any given time interval (such as 5 minutes) can be used for runtime calibration.

\begin{figure}[tbp]
	\centering        
        \includegraphics[width=0.498\textwidth]{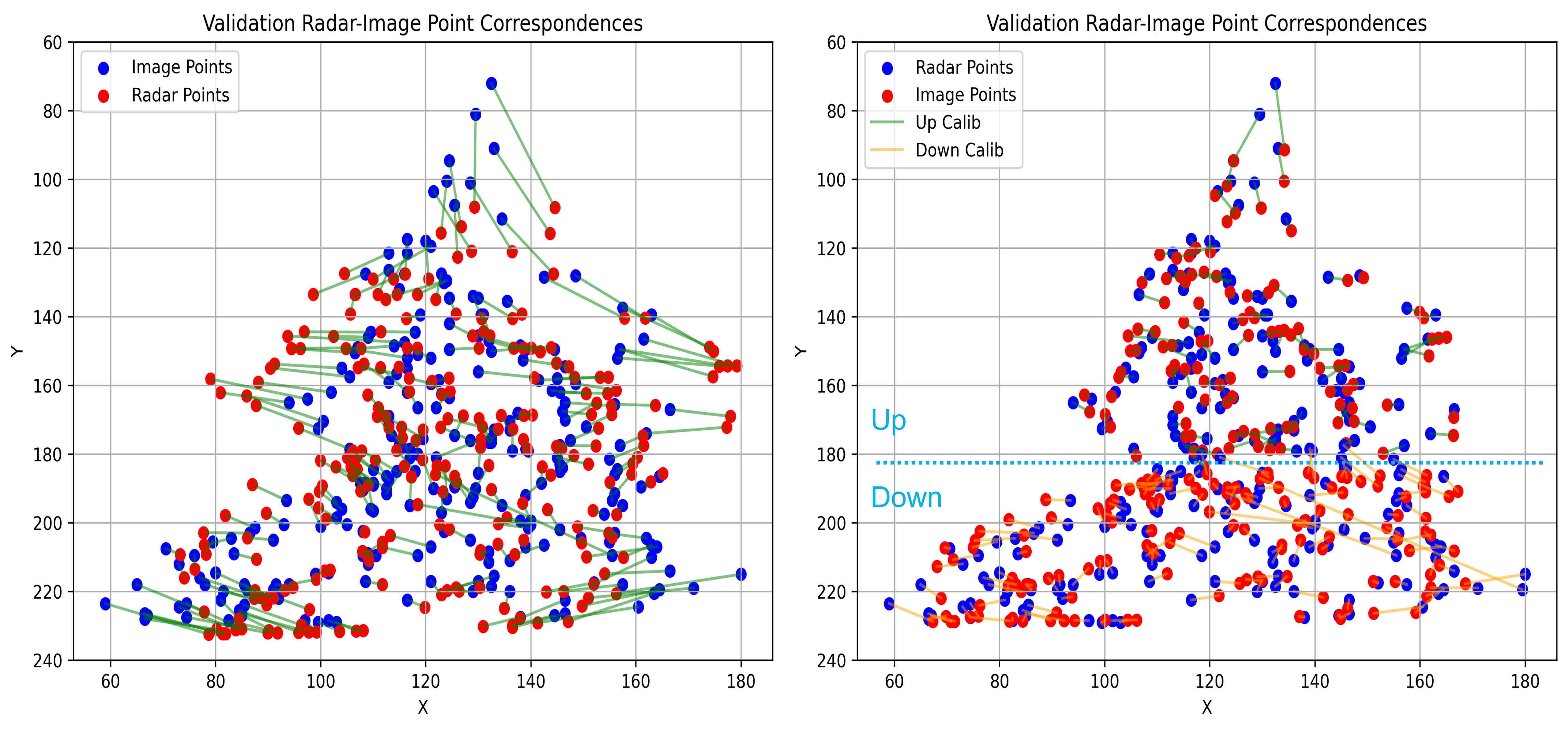}
	\caption{Calibration Results Comparison: Without (left) and With (right) Up-Down Separation Calibration.} 
	\label{sepa}
\end{figure}

Finally, we propose the Up-Down Separation Calibration method to address the discrepancies in sensor measurement accuracy between proximal and distal targets. The point correspondence calibration assumes detected targets are point targets. However, proximal targets are more challenging to be considered as point targets compared to distal ones, and it is known that cameras are more prone to distortion with distant targets. Consequently, indiscriminately using both proximal and distal targets for calibration will result in compromised accuracy. 
To mitigate this, we divide the sensor's field of view into upper and lower sections, corresponding to distant and proximal regions respectively, and perform independent calibrations for each section. Specifically, we partition the camera image based on its height \( H_{img}\), with the upper section defined as \(v < \frac{2}{3} H_{img} \) and the lower section as \(v \geq \frac{2}{3} H_{img} \). This approach allows us to apply specialized calibration matrices for targets in each region, thereby achieving higher accuracy compared to a unified calibration approach, as illustrated in Fig. \ref{sepa} and Table \ref{tab6}. The complete radar-camera online calibration procedure is presented in Algorithm \ref{alg:onlinecalib}.

\subsection{Data Processing in Radar Branch of MOT}\label{AA}
Adhering to the tracking-by-detection paradigm, we first use RadarYOLO to perform object detection on the radar data. As previously discussed, the radar point cloud data and the clustering algorithms applied to it face challenges in radar object detection performance. Hence, we chose to work with low-level radar data, specifically the raw RAD data cube, to leverage the rich information contained in the RAD data. Additionally, since radar object detection focuses more on localizing targets rather than determining their velocity, we opted to ignore the Doppler dimension and use the RA heatmap as input. 
The RA heatmap, a BEV representation created by summing intensity across the Doppler dimension of RAD data, plots azimuth (angle) on the x-axis and range (distance) on the y-axis. This BEV representation can be processed as image data, facilitating the use of established computer vision techniques. Therefore, we adapt the YOLOv4 \cite{bochkovskiy2020yolov4} model to construct RadarYOLO, as illustrated in Fig. \ref{com}, as our radar detector, thereby leveraging YOLO's powerful object detection capabilities to achieve accurate radar object detection. When RadarYOLO processes an RA map, it outputs bounding boxes for the detected objects. The centers of these boxes indicate the objects' positions $P_r^{RA}$ in the radar's RA polar coordinate system. These polar coordinates can then be transformed into a 2D Cartesian radar coordinate system $(x, y)$ using the following equations:
\begin{equation}
\label{eq9}
%\begin{split}
x = r \cdot \sin(\theta), \quad
y = r \cdot \cos(\theta).
%\end{split}
\end{equation}

\subsection{Data Processing in Camera Branch of MOT}
For the camera branch, we use YOLOv4 as the object detector, which can process each image frame to yield a list of category labels and bounding boxes for all detected objects. The position of each detection target within the image is inferred from the lower center point of these bounding boxes. However, unlike radar points that directly represent spatial locations in the physical world, these image-detected points lack such real-world applicability. Hence, transforming these positions into physically meaningful locations is essential. 
Fortunately, using our online targetless calibration, we can easily deduce the physically meaningful locations of camera-detected objects by projecting their positions from the camera plane onto the radar RA plane. Specifically, for any camera-detected object position \( P_p = (u, v) \), we can determine its position \( \hat{P_p} = (r_p, \theta_p) \) within the radar coordinate system using the Homography calibration transformation (see Eq. \ref{eq_H} ):
\begin{equation}
r_p = \frac{h_{11}u + h_{12}v + h_{13}}{h_{31}u + h_{32}v + h_{33}}, \quad \theta_p = \frac{h_{21}u + h_{22}v + h_{23}}{h_{31}u + h_{32}v + h_{33}},
\end{equation}
and further determine its real-world ground position by using Eq. \ref{eq9}:
\begin{equation}\label{eq91}
x_p = r_p \cdot \sin(\theta_p), \quad
y_p = r_p \cdot \cos(\theta_p).
\end{equation}
Additionally, this calibration process also enables us to establish a direct link between the objects detected by the camera and those detected by the radar, thereby significantly aiding sensor fusion.

\begin{table*}[htbp]
\centering
\caption{CLEAR-MOT METRICS FOR TRACKING PERFORMANCE COMPARISON FOR THE CONTROLLED EXPERIMENTS (95\% CONFIDENCE INTERVALS FROM 5 TRIALS)}
\label{tab4}
\resizebox{\textwidth}{!}{
\begin{tabular}{@{}l l c c c c c@{}} 
\toprule
\textbf{Scenario} & \textbf{Tracker} & \textbf{FPR (\%)} & \textbf{FNR (\%)} & \textbf{IDSWR (\%)} & \textbf{MOTA (\%)} & \textbf{MOTP (m)} \\
\midrule
\multirow{3}{*}{Scenario 1}
 & Camera Tracker & $0.00\pm0.00$ & $3.65\pm0.41$ & $0.00\pm0.00$ & $96.35\pm0.41$ & $1.36\pm0.12$ \\
 & Radar Tracker & $0.03\pm0.01$ & $21.91\pm1.55$ & $0.18\pm0.05$ & $77.88\pm1.61$ & $0.57\pm0.05$ \\ 
 & Sensor Fusion Tracker & $3.83\pm0.37$ & $0.00\pm0.00$ & $0.18\pm0.04$ & $95.99\pm0.39$ & $0.88\pm0.07$ \\ 
\midrule
\multirow{3}{*}{Scenario 2}
 & Camera Tracker & $0.01\pm0.01$ & $4.00\pm0.48$ & $0.01\pm0.01$ & $95.98\pm0.46$ & $1.40\pm0.14$ \\
 & Radar Tracker & $0.04\pm0.01$ & $22.50\pm1.74$ & $0.20\pm0.05$ & $77.26\pm1.80$ & $0.52\pm0.06$ \\ 
 & Sensor Fusion Tracker & $3.99\pm0.42$ & $0.00\pm0.00$ & $0.18\pm0.04$ & $95.83\pm0.43$ & $0.85\pm0.06$ \\ 
\midrule
\multirow{3}{*}{Scenario 3}
 & Camera Tracker & $0.00\pm0.00$ & $3.75\pm0.45$ & $0.00\pm0.00$ & $96.25\pm0.45$ & $1.74\pm0.16$ \\
 & Radar Tracker & $0.02\pm0.01$ & $22.00\pm1.69$ & $0.19\pm0.05$ & $77.79\pm1.72$ & $0.58\pm0.05$ \\ 
 & Sensor Fusion Tracker & $3.70\pm0.39$ & $0.00\pm0.00$ & $0.20\pm0.05$ & $96.10\pm0.41$ & $0.91\pm0.07$ \\ 
\bottomrule
\end{tabular}}
\end{table*}

\begin{table*}[htbp]
\centering
\caption{FNR AND MOTP COMPARISON BETWEEN THE TRACKING OF PERSON AND CAR FOR THE CONTROLLED EXPERIMENTS (95\% CONFIDENCE INTERVALS FROM 5 TRIALS)}
\label{tab5}
\resizebox{\textwidth}{!}{
\begin{tabular}{@{}l c cc cc cc@{}} 
\toprule
\multirow{2}{*}{\textbf{Scenario}} & \multirow{2}{*}{\textbf{Target}} & \multicolumn{2}{c}{\textbf{Camera Tracker}} & \multicolumn{2}{c}{\textbf{Radar Tracker}} & \multicolumn{2}{c}{\textbf{Sensor Fusion Tracker}} \\
\cmidrule(lr){3-4} \cmidrule(lr){5-6} \cmidrule(l){7-8}
 & & FNR (\%) & MOTP (m) & FNR (\%) & MOTP (m) & FNR (\%) & MOTP (m) \\
\midrule
\multirow{2}{*}{Scenario 1} 
 & Person & $3.65\pm0.41$ & $1.83\pm0.14$ & $21.68\pm1.58$ & $0.37\pm0.04$ & $0.00\pm0.00$ & $1.12\pm0.08$ \\
 & Car    & $0.00\pm0.00$ & $0.89\pm0.08$ & $0.23\pm0.04$ & $0.77\pm0.05$ & $0.00\pm0.00$ & $0.64\pm0.06$ \\
\midrule
\multirow{2}{*}{Scenario 2}
 & Person & $4.00\pm0.48$ & $1.85\pm0.15$ & $22.20\pm1.62$ & $0.41\pm0.04$ & $0.00\pm0.00$ & $1.15\pm0.09$ \\
 & Car    & $0.00\pm0.00$ & $0.94\pm0.09$ & $0.30\pm0.05$ & $0.63\pm0.06$ & $0.00\pm0.00$ & $0.55\pm0.05$ \\
\midrule
\multirow{2}{*}{Scenario 3}
 & Person & $3.75\pm0.45$ & $1.80\pm0.14$ & $21.90\pm1.59$ & $0.34\pm0.04$ & $0.00\pm0.00$ & $1.10\pm0.08$ \\
 & Car    & $0.00\pm0.00$ & $1.68\pm0.17$ & $0.10\pm0.02$ & $0.82\pm0.07$ & $0.00\pm0.00$ & $0.72\pm0.06$ \\
\bottomrule
\end{tabular}}
\end{table*}

\begin{table*}[htbp]
\centering
\caption{CLEAR-MOT METRICS FOR TRACKING PERFORMANCE COMPARISON FOR THE REAL-WORLD EXPERIMENTS (95\% CONFIDENCE INTERVALS FROM 5 TRIALS)}
\label{tab_realworld}
\resizebox{\textwidth}{!}{
\begin{tabular}{@{}l l c c c c c@{}} 
\toprule
\textbf{Scenario} & \textbf{Tracker} & \textbf{FPR (\%)} & \textbf{FNR (\%)} & \textbf{IDSWR (\%)} & \textbf{MOTA (\%)} & \textbf{MOTP (m)} \\
\midrule
\multirow{3}{*}{Campus Street}
 & Camera Tracker & $0.02\pm0.01$ & $5.10\pm0.55$ & $0.05\pm0.02$ & $94.83\pm0.58$ & $1.41\pm0.13$ \\
 & Radar Tracker & $0.04\pm0.01$ & $29.75\pm2.05$ & $0.10\pm0.03$ & $70.11\pm2.12$ & $0.60\pm0.05$ \\ 
 & Sensor Fusion Tracker & $4.20\pm0.40$ & $0.25\pm0.10$ & $0.08\pm0.03$ & $95.47\pm0.42$ & $0.89\pm0.07$ \\ 
\midrule
\multirow{3}{*}{Urban Intersection}
 & Camera Tracker & $0.03\pm0.01$ & $6.20\pm0.62$ & $0.07\pm0.02$ & $93.70\pm0.65$ & $1.52\pm0.14$ \\
 & Radar Tracker & $0.05\pm0.01$ & $31.65\pm2.20$ & $0.12\pm0.04$ & $68.18\pm2.30$ & $0.62\pm0.05$ \\ 
 & Sensor Fusion Tracker & $4.45\pm0.43$ & $0.30\pm0.12$ & $0.09\pm0.03$ & $95.16\pm0.45$ & $0.91\pm0.08$ \\ 
\bottomrule
\end{tabular}}
\end{table*}

\begin{table}[htbp]
\centering
\caption{REPROJECTION ERRORS COMPARISON OF RADAR-CAMERA MUTUAL CALIBRATION RESULTS, AND USING AND WITHOUT USING UP-DOWN SEPARATION CALIBRATION}
\label{tab6}
\begin{adjustbox}{width=0.48\textwidth}
\begin{tabular}{@{}lcccc@{}} 
\toprule
& \multicolumn{2}{c}{\textbf{$\mathcal{E}_{rep}$ of RadToCam}} & \multicolumn{2}{c}{\textbf{$\mathcal{E}_{rep}$ of CamToRad}} \\
\cmidrule(r){2-3} \cmidrule(l){4-5} 
 & Not Up-Down  & Up-Down & Not Up-Down  & Up-Down \\
\midrule
Scenario 1 & 73.02 & 52.79 & 22.57 & 15.42 \\
Scenario 2 & 87.18 & 43.27 & 23.44 & 17.72 \\
Scenario 3 & 88.03 & 44.28 & 19.42 & 12.19 \\
\bottomrule
\end{tabular}
\end{adjustbox}
\end{table}

\begin{figure*}[h!]
	\centering        
        \includegraphics[width=0.98\textwidth]{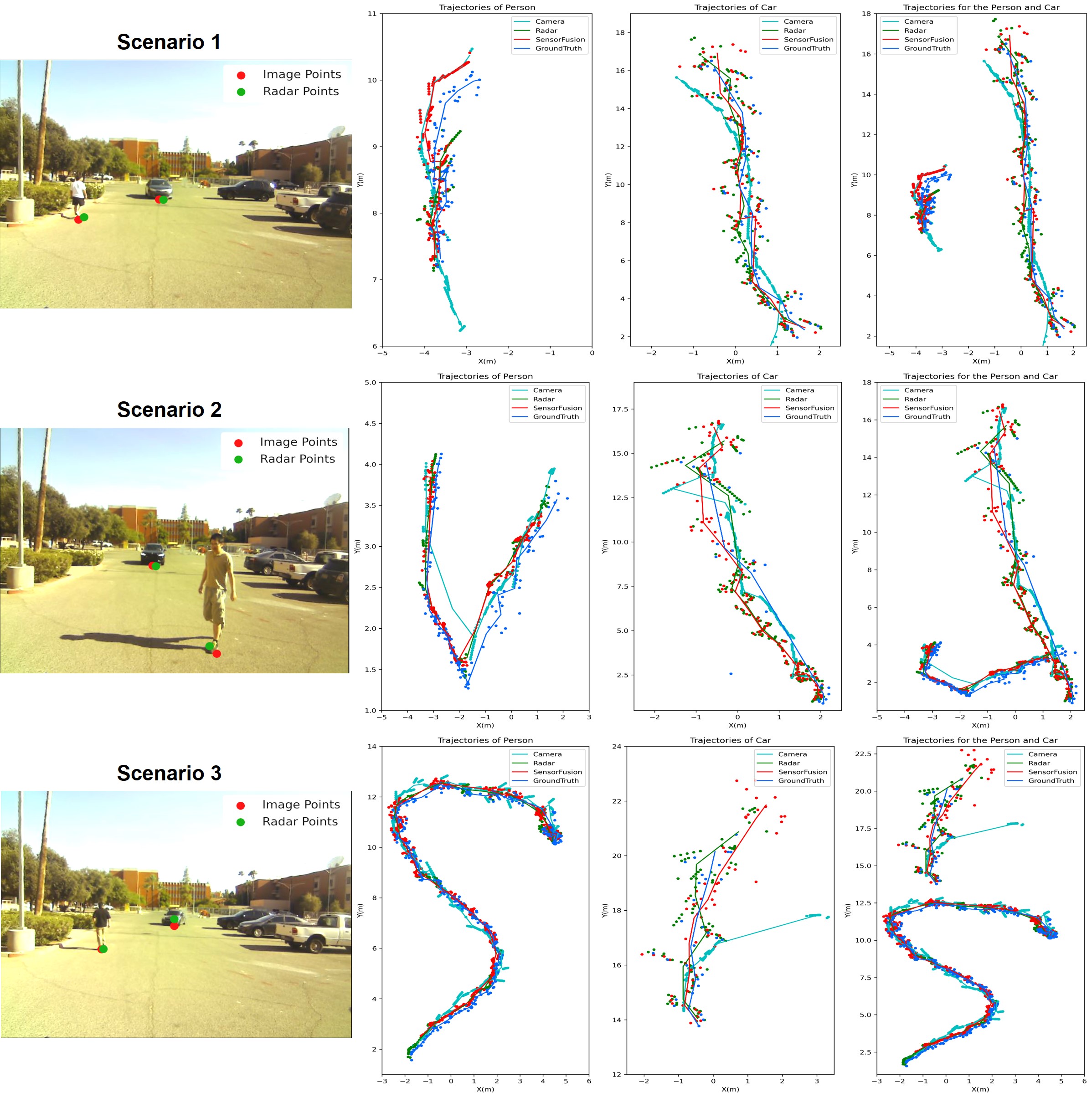}
	\caption{Calibration and Trajectory Results for Person and Car Across Three Scenarios using the Proposed Method}
	\label{5}
\end{figure*}

\subsection{Data Processing in Sensor Fusion Branch of MOT}
Sensor fusion tracking aims to integrate radar and camera sensor data, to achieve more robust and accurate tracking performance. To effectively fuse the detections from both sensors, we employ a two-stage matching strategy: feature matching followed by position matching. In the feature matching stage, the Common Feature Discriminator assesses the similarity between all detected objects in a radar frame and its temporally-aligned image frame. The feature matching confidence score (representing the discriminator’s binary classification confidence), \( \bm{\mathcal{S}}(f_r, f_c) \), is computed for each object pair. A match is considered successful if:
\begin{equation}
%\small
\begin{aligned}
\bm{\mathcal{S}}(f_r, f_c) > 0.8 \quad \text{and} \quad \| P_r - P_p \| \leq 3 \, \text{meters}, 
\end{aligned}
\end{equation}
where \( f_r \) and \( f_c \) are feature vectors from a pair of radar and camera objects, \( P_r \) and \( P_p \) are their positional coordinates, and \( \| P_r - P_p \| \) is the Euclidean distance between them.
For objects not matched in the feature matching stage, position matching is employed. This involves creating a distance cost matrix \( M \) for radar and camera detections, where each element \( M_{ij} \) is the Euclidean distance between the \( i^{th} \) radar object and the \( j^{th} \) camera object:
\begin{equation}
%\small
\begin{aligned}
M_{ij} = \| P_{r_i} - P_{p_j} \| ,
\end{aligned}
\end{equation}
The Hungarian algorithm is applied to the matrix \( M \) to find the optimal matching pairs. In this stage, any pair with \( M_{ij} > 3 \, \text{meters} \) is considered unmatched.

Following the completion of the two-stage matching process, objects that are successfully matched are processed in a manner that capitalizes on the respective strengths of each sensor. Specifically, the sensor fusion positions  \( P_{\text{sf}} \) for these objects are determined based on the radar detection positions due to radar's superior positional precision. Concurrently, their categorical information \( C_{\text{sf}} \) is derived from the camera detection's category data \( C_c \), utilizing the camera's enhanced accuracy in classification. For objects that are not successfully matched, the sensor fusion positions default to their original sensor detection coordinates. In these instances, unmatched radar detections are devoid of category information, whereas unmatched camera detections maintain their pre-existing category classifications. That is:
\begin{equation}
%\small
\begin{aligned}
(P_{\text{sf}},\ C_{\text{sf}})= \begin{cases} 
(P_{\text{r}},\ C_{\text{c}}), & \text{if matched objects} \\
(P_{\text{p}},\ C_{\text{c}}), & \text{if camera unmatched} \\
(P_{\text{r}},\ -), & \text{if radar unmatched}
\end{cases}.
\end{aligned}
\end{equation}

Additionally, in the process of matching current detections \( D \) with established tracks \( T \), a category-consistency check strategy is employed:
\begin{equation}
%\small
\begin{aligned}
\displaystyle \bm{\mathcal{F}}_{\text{consistency}}(C_{D}, C_{T}) = 
\begin{cases} 
\text{Matched}, & \text{if } C_{D} = C_{T} \\
\text{Unmatched}, & \text{if } C_{D} \neq C_{T}
\end{cases},
\end{aligned}
\end{equation}
where \( C_{D} \) and \( C_{T} \) represent the categories of current detections and existing tracks, respectively.
This strategy involves a meticulous examination of the object categories associated with both current detections and existing tracks. By doing so, we effectively exclude pairs of detections and tracks that exhibit category mismatches, thereby circumventing erroneous associations of an object with another object's track—a frequent issue in traditional detection-track matching methodologies that rely only on positional data. This incorporation of a category-consistency check thus significantly enhances the precision of the detection-track matching procedure within the sensor fusion framework.

\section{Experiments And Results}
\subsection{Experimental Setup}
\begin{figure}[htbp]
	\centering        
        \includegraphics[width=0.45\textwidth]{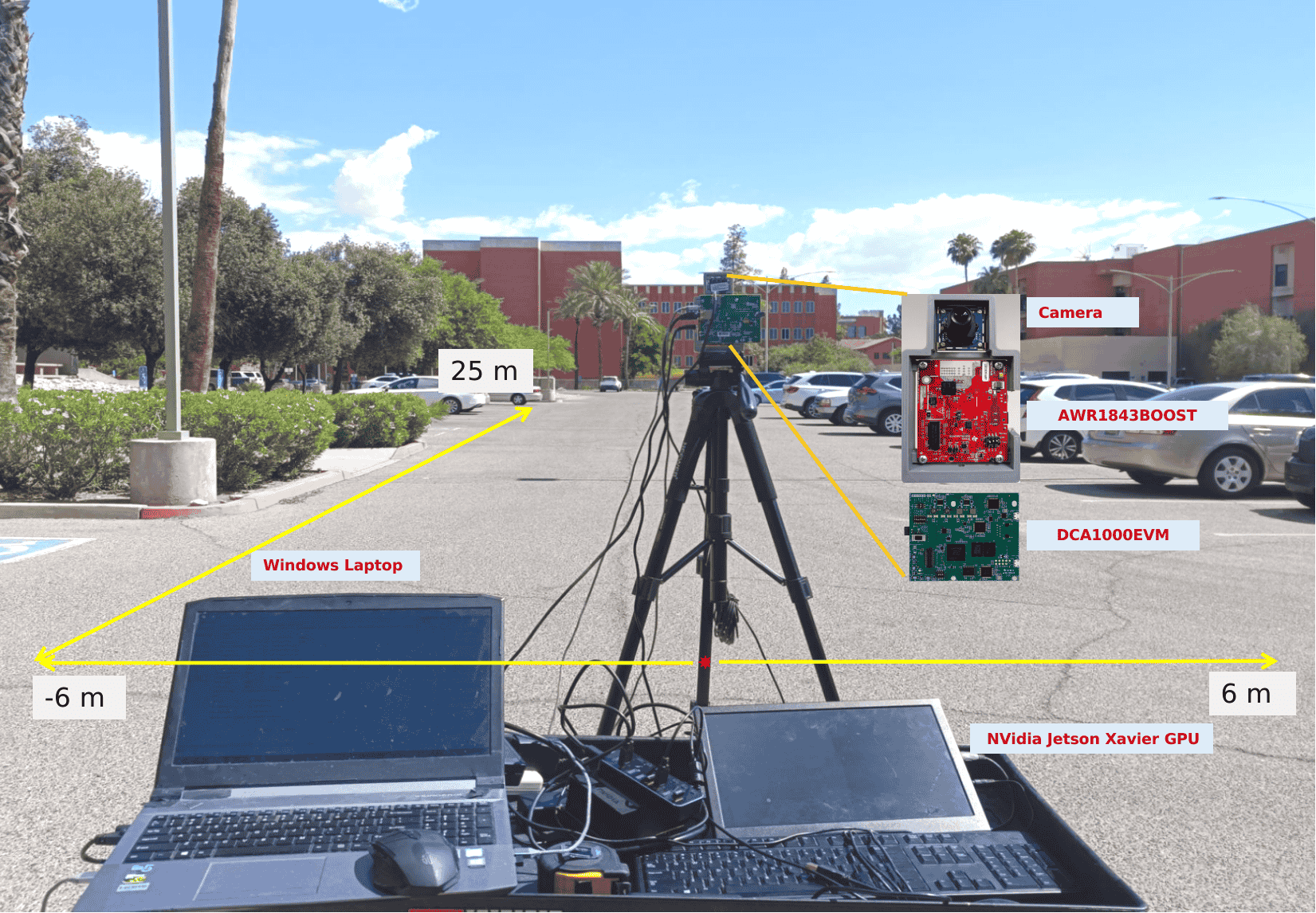}
	\caption{Data Collection Setup: A Windows laptop and an NVidia Jetson Xavier establish a connection with the radar-camera system using Ethernet and USB cables to collect data. } 
	\label{3}
\end{figure}

Our experiment uses two main sensing modalities: a TI AWR1843BOOST mmWave radar transceiver with antenna channels in azimuth and elevation for 3D sensing, and a USB8MP02G monocular HD digital camera for high-definition image captures. To acquire raw radar data, we utilized a TI DCA1000 evaluation module (EVM). These sensors were secured to a static tripod mounted using a 3D printed frame.
For image capture, we employed an Nvidia Jetson Xavier running Ubuntu 18 and ROS Melodic, which launches YOLO to capture images from the camera. Meanwhile, a Windows laptop was used to collect radar raw data by operating the mmWave studio GUI tool. The experimental setup is shown in Fig. \ref{3}.

\subsection{Performance Evaluation}
We used the CLEAR metrics \cite{bernardin2008evaluating} to evaluate our tracking approach. These metrics are $MOTA$ (Accuracy) and $MOTP$ (Precision).
$MOTA$ is the ratio of correctly tracked objects to all objects in the scene. It is computed as follows:
\begin{equation}
\small
\begin{aligned}
MOTA &= 1 - \frac{\sum_{t=1}^{T}({FN_t} + {FP_t} + {IDSW_t})}{\sum_{t=1}^{T}{GT_t}} ,\\
&= 1 - FNR  - FPR  - IDSWR,
\end{aligned}
\end{equation}
where $T$ is the number of frames, ${GT_t}$ is the number of real objects, ${FN_t}$ is the number of missed objects, ${FP_t}$ is the number of spurious objects, and ${IDSW_t}$ is the number of id switches in the $t^{th}$ frame. The ${FNR}$, ${FPR}$, and ${IDSWR}$ are the rates of false negatives, false positives, and id switches, respectively.

$MOTP$ is the average distance between the tracked and ground truth positions of each object. It is calculated as follows:
\begin{equation}
MOTP = \frac{\sum_{t=1}^{T} \sum_{i=1}^{N} d_t^i}{\sum_{t=1}^{T}{GT_t}},
\end{equation}
where $N$ is the number of objects in the scene, and $d_t^i$ is the distance between the tracked and ground truth positions of the $i^{th}$ object in the $t^{th}$ frame. In this paper, we employ the radar-sensor fusion method from our previous work \cite{sengupta2022robust}, utilizing fine-tuned camera BEV transformation and incorporating specific manual corrections (based on pre-measured key points in the experimental site) to generate these ground truth positions.

\begin{figure*}[h!]
	\centering        
        \includegraphics[width=0.92\textwidth]{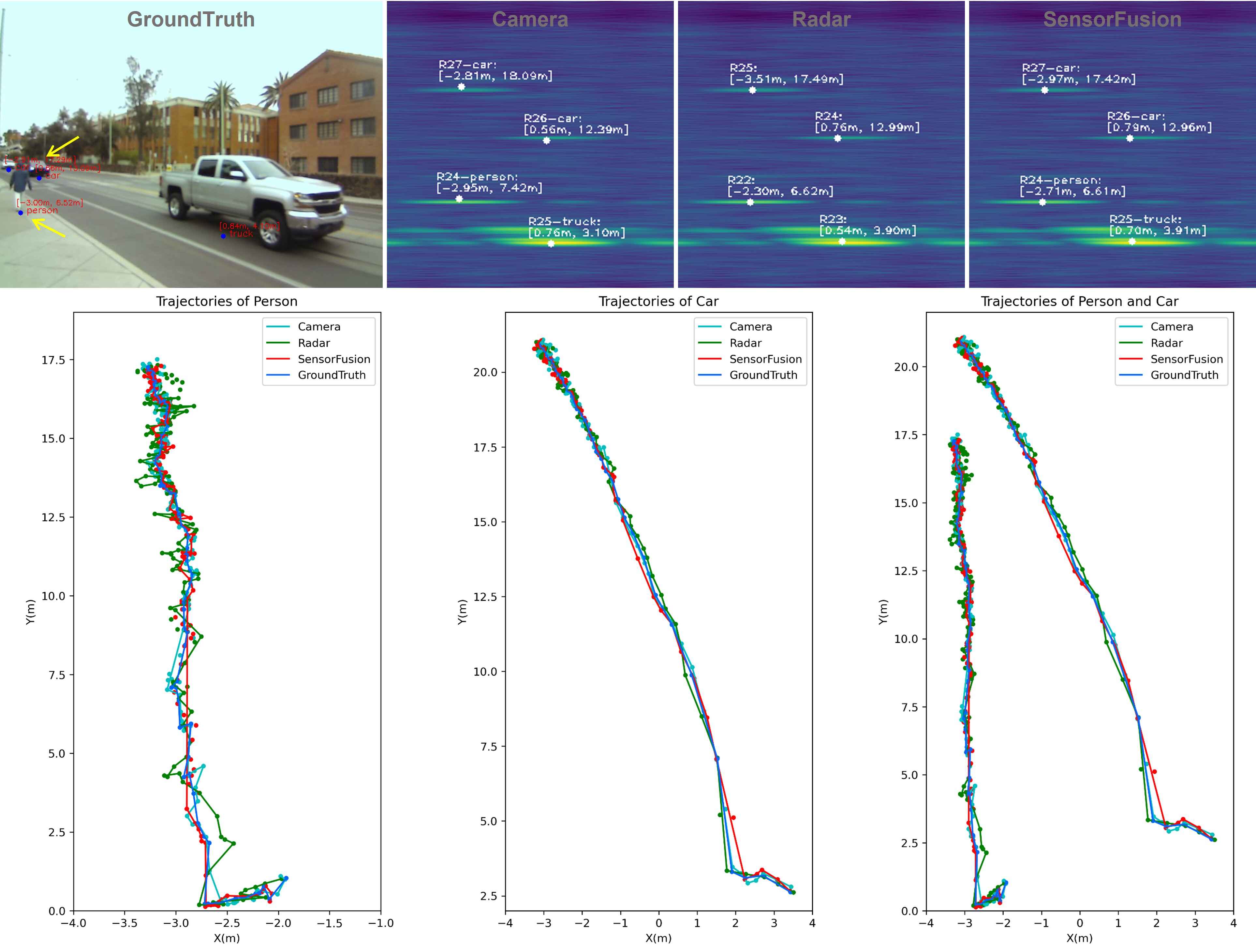}
	\caption{Tracking and Trajectory Results of the Proposed MOT Method in the Campus Street under Overcast Weather.}
	\label{street}
\end{figure*}

\begin{figure*}[h!]
	\centering        
        \includegraphics[width=0.91\textwidth]{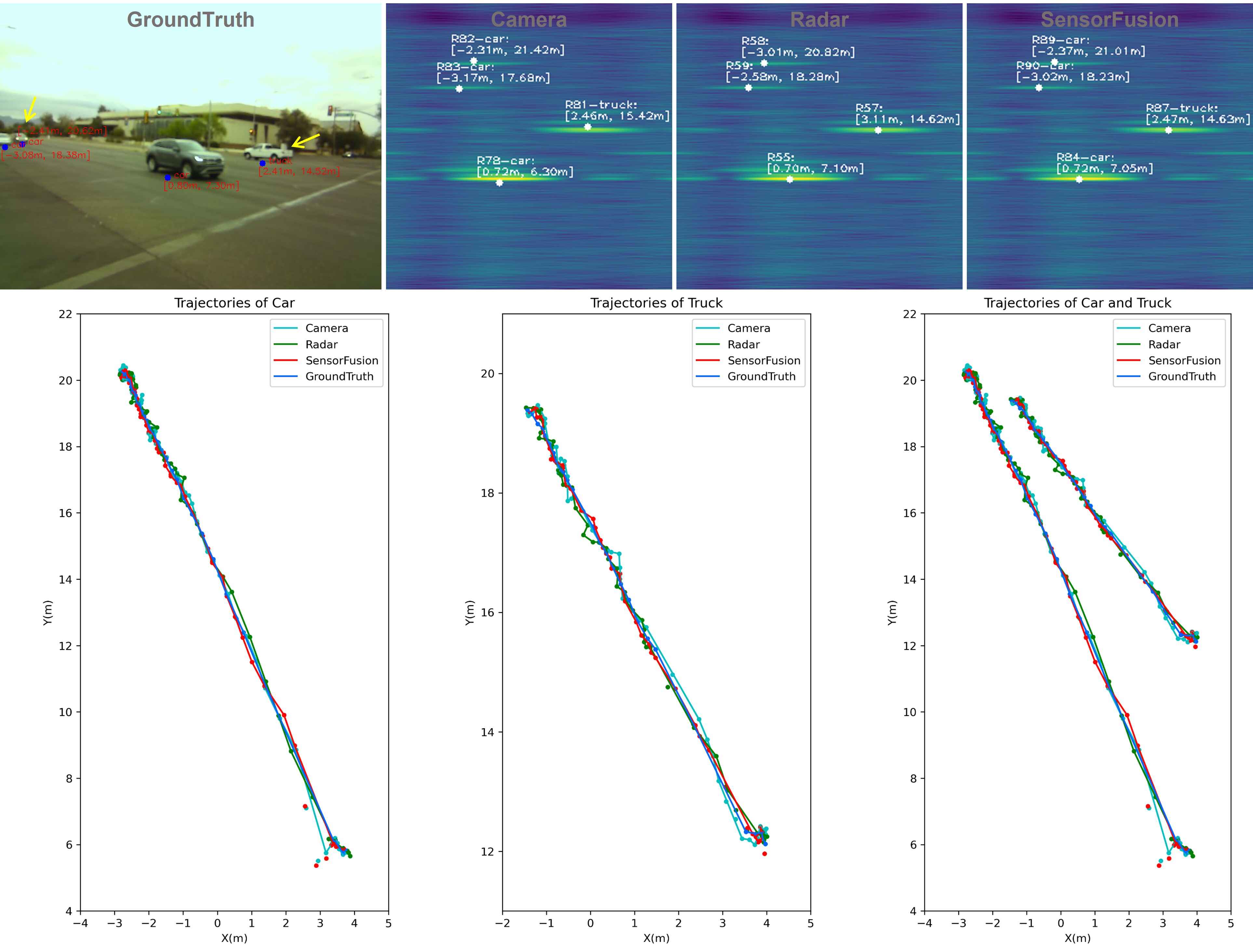}
	\caption{Tracking and Trajectory Results of the Proposed MOT Method in a Busy Urban Intersection under Overcast Weather.}
	\label{road}
\end{figure*}

\begin{figure*}[h!]
	\centering        
        \includegraphics[width=0.98\textwidth]{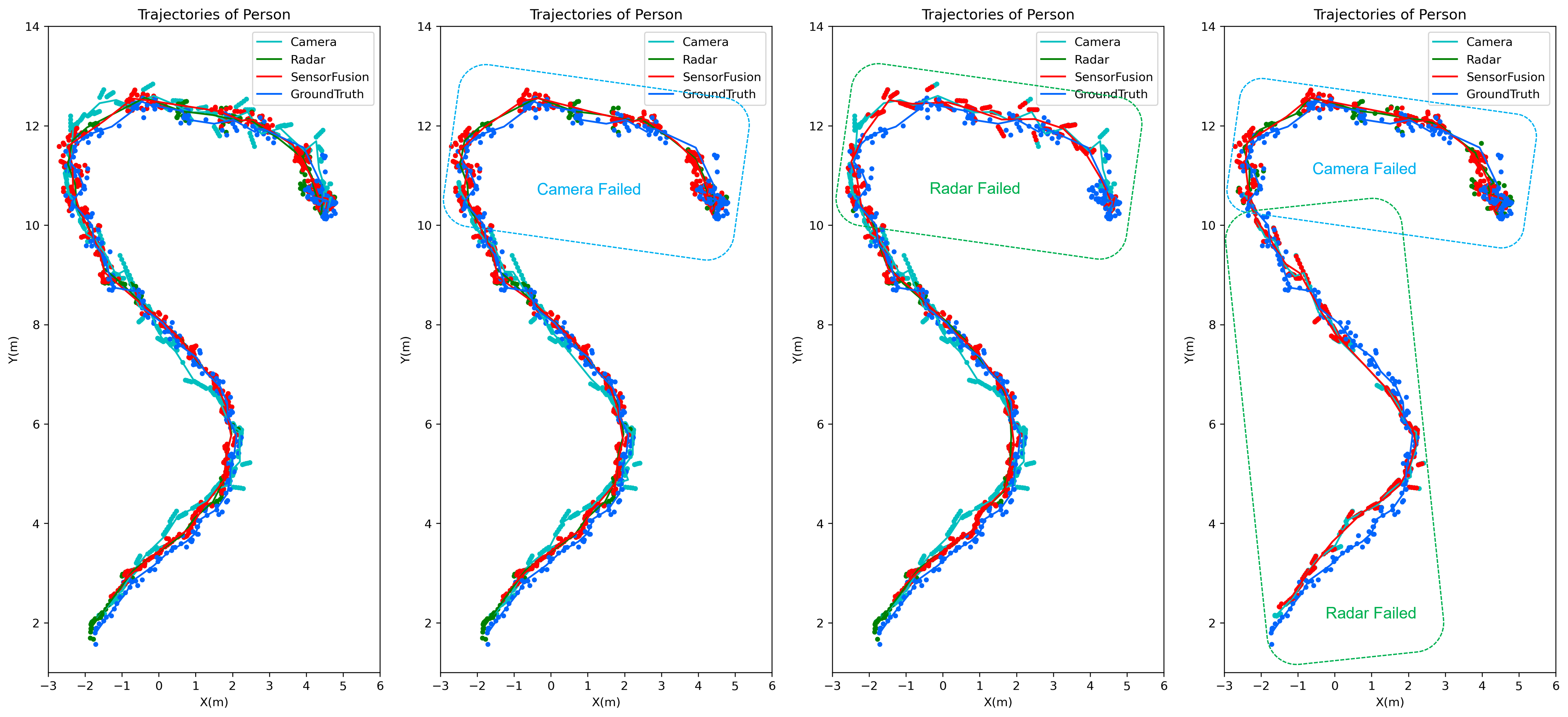}
	\caption{Sensor Fusion Results for Person in Scenario 3 with Camera, Radar, and Radar-then-Camera Failed, Respectively.}
	\label{fail}
\end{figure*}

\subsection{Results and Discussion}
To effectively evaluate the performance of our proposed MOT method, we conducted experiments in both controlled and real-world environments. The controlled experiments were carried out in a parking lot under clear weather conditions and included three different scenarios, each involving different combinations of persons and cars, as shown in Fig. \ref{5}. For the real-world experiments, we selected two actual traffic environments: a campus street (Fig. \ref{street}) and a busy urban intersection (Fig. \ref{road}), both under overcast weather with light rain. To ensure statistical reliability, the proposed MOT method was executed five times on the data from each experiment, and the averaged tracking performance along with corresponding 95\% confidence intervals are reported in Tables \ref{tab4}, \ref{tab5}, and \ref{tab_realworld}.

In the controlled experiments, both the persons and the cars were allowed to move freely within the designated area (measuring 25 meters long and 12 meters wide) while data was collected.
The experimental results, as detailed in Table \ref{tab4}, indicate that our radar-camera fusion framework for MOT demonstrates more reliable and robust performance compared to the standalone camera and radar trackers. For scenario 1, the sensor fusion tracker achieves a remarkably low FNR of 0\%, a significant improvement over the standalone systems. The radar tracker, in particular, suffers from a high FNR of 21.91\%, largely due to its limited angular resolution, which hampers its ability to differentiate between closely located objects (as shown in Frame 109 of Fig. \ref{4}). On the other hand, when vehicles approach the radar, their large size and strong reflective nature might overshadow the radar reflections from pedestrians (as shown in Frame 200 of Fig. \ref{4}), contributing to higher FNR. The camera tracker fares better with an FNR of only 3.65\%, primarily because occlusions (as shown in Frame 261 of Fig. \ref{4}), which are the main source of missed detections, occur less frequently in our experimental setup involving only two objects. By integrating inputs from both the camera and radar, the Sensor Fusion Tracker excels in minimizing missed detections, thereby significantly enhancing overall detection and tracking comprehensiveness, as demonstrated in Fig. \ref{4}. Similar results can also be observed for scenarios 2 and 3. 

Furthermore, in terms of MOTA and MOTP, the sensor fusion tracker achieves a balanced performance. Specifically, it has a lower MOTA than the camera tracker but higher than the radar tracker, while for MOTP, it is the opposite. The reasons are evident. Firstly, radar can accurately measure object distances, thus providing high MOTP. However, radar suffers from significant detection noise, leading to more false positives. Its low data resolution also hampers effective object identification, resulting in more IDSW and consequently lower MOTA. On the other hand, the camera’s high resolution reduces false negatives and IDSW, resulting in high MOTA. However, the camera cannot accurately measure the true distances of objects, and the distance derivation using camera-to-radar calibration is subject to calibration errors, leading to poor MOTP. As shown in Fig. \ref{4}, many camera-detected positions after projection by using the calibration matrix do not fall well on the radar detection center. The sensor fusion tracker, by integrating camera data, achieves better MOTA (as shown in Table \ref{tab4}, close to that of the camera tracker), while integrating radar data results in better MOTP. Nevertheless, compared to the standalone trackers, the sensor fusion tracker also has drawbacks. The fusion process that combines detections from both the camera and radar can potentially generate spurious tracks due to discrepancies in detection accuracy between the two sensors, thereby exhibiting higher FPR and IDSW, which leads to a lower MOTA. This can also cause incorrect target associations, further degrading MOTP.

Table \ref{tab5} compares tracking performance for the person and car, revealing that the camera tracker achieves higher tracking precision for cars, while the radar tracker achieves higher precision for persons but also produces significantly more false negatives (missed detections) for persons. This discrepancy can be attributed to several factors. For the camera tracker, the larger size of cars leads to more stable detection, fewer false negatives, and more compact bounding boxes, resulting in more accurate position derivation. Additionally, cars move at consistent speeds and follow more linear paths, making them easier to track compared to pedestrians. However, despite the large size and high reflectivity of cars making them less prone to false negatives for the radar tracker, this attribute can also be a drawback. Radar reflections can originate from any part of the car's body, not necessarily the front, causing significant jitter in the detected position of the car. This ultimately leads to less precise car position measurements. Notably, due to the weaker reflectivity of persons, their detections can be overshadowed by high-reflectivity objects like cars \cite{9667163}, resulting in many missed detections.

In addition to the quantitative results, we also provide the trajectories of the two objects (person and car) generated by the proposed method, as shown in Fig. \ref{5}. It is easy to see from the figure that the trajectory prediction of sensor fusion is better. For the person trajectory, the camera tracker may deviate from the real trajectory due to calibration errors and non-compact bounding boxes. The radar tracker, on the other hand, cannot track the person continuously due to inaccurate classification or missed detections. Only the sensor fusion tracker can continuously track the object and generate a relatively accurate trajectory. For the car trajectory, the camera tracker has the worst predicted trajectory for the same reasons as above. The radar tracker predicts trajectories more accurately thanks to its precise object localization. The sensor fusion tracker, by integrating inputs from the camera tracker, compensates for radar's inaccurate classification and missed detections and achieves the most accurate results. In addition, the sensor fusion tracker has the advantage of integrating the input from the camera tracker. On the one hand, using the object category information provided by the camera tracker can prevent erroneous matching during trajectory generation. For example, as shown in Frame 139 of Fig. \ref{4}, a spurious object detected by the radar leads to a spurious person object in the sensor fusion tracker. However, the spurious person object will be deleted after being retained for a certain period of time because it cannot find a matching object (of the same category and with a close distance) in the subsequent matching. By Frame 152, this error has been removed. On the other hand, by using image features to perform radar-camera feature matching, it can avoid directly matching two objects at close distances. For example, in Frame 139, the camera tracker's car object will not be matched with the spurious object in the radar tracker, even though they are close in distance. Instead, it will be matched with another object because their features are more similar.

Similar experimental results were observed in our real-world experiments, despite the fact that these two dynamic traffic environments occurred under overcast weather conditions with light rain, involved more objects, and reflected more realistic and complex motion patterns. As detailed in Table \ref{tab_realworld}, these scenarios exhibited trends consistent with those from the controlled environment. Specifically, radar’s detection performance remained robust against weather variations, showing minimal influence from rain or overcast skies. Radar consistently detected distant moving objects at intersections without significant interference from road clutter, which was evidenced by its relatively stable MOTP values (approximately $0.60$m and $0.62$m, respectively) across both real-world scenarios. Conversely, camera detections became slightly blurred under rainy conditions, causing the FNR to increase slightly (e.g., $6.20\%$ at the urban intersection, compared to $3.65\%$ in controlled Scenario 1). Nevertheless, the YOLO detector maintained stable tracking performance overall, achieving high MOTA values ($94.83\%$ and $93.70\%$ respectively).
However, radar performance notably degraded in situations involving multiple closely positioned objects near the sensor. Radar frequently struggled to distinguish these objects, likely due to strong overlapping radar reflections from adjacent surfaces, resulting in significant missed detections. This limitation is clearly illustrated by the notably higher radar FNR ($29.75\%$ for the campus street and $31.65\%$ for the urban intersection), substantially higher than the controlled experiments. Despite these radar limitations, the sensor fusion tracker effectively mitigated missed detections by integrating camera detections, maintaining an impressively low FNR (approximately $0.25\%$ and $0.30\%$, respectively). However, this improvement occasionally introduced false positives (FPR around $4.20\%$ and $4.45\%$) due to imperfect radar-camera associations.

That said, several distinctive characteristics emerged when comparing the real-world scenarios to the controlled experiments. Notably, in real-world traffic scenes, objects typically moved faster and adhered strictly to structured routes, such as sidewalks for pedestrians or designated lanes for vehicles. Consequently, object trajectories were generally more linear and less varied, contributing to higher tracking accuracy and significantly lower ID Switch rates (IDSWR around $0.08\%$ and $0.09\%$ for the sensor fusion tracker, compared to controlled experiments around $0.18\%$). Furthermore, radar demonstrated exceptional effectiveness in tracking larger vehicles at greater distances, benefiting from stable and strong radar reflections from the metallic surfaces of vehicles. Nonetheless, radar continued to exhibit considerable difficulties when reliably tracking smaller or closely grouped objects such as pedestrians, highlighting the critical importance of camera data for compensating for these radar shortcomings.

Finally, we evaluated the proposed sensor fusion method's ability to handle sensor failures, as shown in Fig. \ref{fail}. It is evident that our sensor fusion tracker effectively manages such situations to ensure continuous and robust tracking. Firstly, when the camera fails to provide detections, the sensor fusion tracker can still rely on radar data to continue tracking objects. In this case, the category-consistency check strategy is disabled, and the object category information is entirely derived from the track itself. Conversely, when the radar fails, the camera compensates for the detection gap, ensuring that tracking does not cease. This means that as long as at least one sensor is operational during tracking, the sensor fusion tracker can guarantee continuous object tracking. Furthermore, as observed in the figure, if the radar fails and only camera data is relied upon, the tracking precision of the sensor fusion tracker decreases. If only radar data is used, the object category cannot be updated, potentially leading to inaccurate object association and impossible continuous same-object tracking. Therefore, by integrating camera and radar data, the sensor fusion tracker can better ensure tracking accuracy, robustness, and reliability.

\subsubsection*{\textbf{Limitations}}
Despite the promising results, our proposed method still faces certain limitations. First, although we hypothesize that common features exist between radar and camera detections of identical objects, and have developed a Common Feature Discriminator to leverage these features, the nature of these features remains unclear. The visualization and interpretability of such learned common features warrant further investigation in future studies. Second, our approach determines the real-world positions of camera-detected objects by projecting image points onto the radar coordinate system, thus heavily relying on the accuracy of radar-camera calibration. Consequently, calibration errors may directly compromise the localization precision of camera-based detections. Third, although radar RAD data is rich in information, it lacks human readability and exhibits limited angular resolution, frequently resulting in missed detections under frequent occlusion conditions, especially for closely spaced targets. Moreover, RadarYOLO requires retraining for each new radar configuration, and the large size of RAD tensors imposes significant computational load. At last, in terms of runtime efficiency, our current implementation achieves a frame rate of approximately 19 Hz (52 ms per frame), which does not meet the real-time requirements. However, we believe the system could achieve this through further optimizations, such as re-implementing the pipeline in a faster C/C++ environment instead of Python and replacing the backbone networks with lightweight alternatives.

\section{Conclusions}
This paper presents an efficient radar-camera fusion framework for multi-object tracking that overcomes the limitations of traditional methods by automating the calibration process and minimizing manual interventions. By leveraging low-level radar RAD data and deep learning techniques, our framework extracts implicit semantic information from radar data and identifies common features with camera data. These shared features enable online targetless radar-camera calibration, allowing for the direct estimation of real-world positions of camera detections without the need for complex transformations that often require manual operation. Furthermore, our approach goes beyond mere position matching by incorporating feature matching, resulting in a more accurate and robust fusion of radar and camera detections. To enhance inter-frame object association accuracy, a category-consistency checking strategy is employed to associate objects across consecutive frames. 

In controlled tests, our Sensor Fusion Tracker achieved near-zero missed detections—dramatically lower than either camera- or radar-only methods—while matching the camera’s identity consistency and the radar’s distance accuracy. In real-world traffic, fusion sustained robust detection and track continuity under challenging conditions, producing the most consistent and accurate trajectories.
The method also proved resilient to intermittent sensor failures—relying on the remaining modality with minimal performance loss. These results confirm that deep‐feature‐driven, targetless calibration with dual‐cue fusion enhances MOT robustness and accuracy. Future work will focus on interpreting learned features, enabling adaptive calibration, and optimizing real-time performance.

%%%%%%%%%%%%%%%%%%%%%%%%%%%%%%%%%%%%%%%%%%%%%

\vspace{0.5cm}

% \begin{thebibliography} environment in LaTeX is used to manually create a bibliography without relying on a .bib file, and it should not be combined with the \bibliography and \bibliographystyle commands.
\bibliographystyle{IEEEtran}
{\small
\bibliography{references.bib}
}

%\end{thebibliography}

% \newpage
\vspace{-20pt}

\vfill

\end{document}